%% file: main.tex
\documentclass{applemlr}

\title{\method: Unlocking Parallel Training of\\
Nonlinear RNNs for Large Language Models}

\input{apple_template/apple_preamble}

\author{Federico Danieli}
\author{Pau Rodr\'iguez}
\author{Miguel Sarabia}
\author{Xavier Suau}
\author{Luca Zappella}

\affiliation{Apple}

\abstract{
Recurrent Neural Networks (RNNs) laid the foundation for sequence modeling, but their intrinsic sequential nature restricts parallel computation, creating a fundamental barrier to scaling. This has led to the dominance of parallelizable architectures like Transformers and, more recently, State Space Models (SSMs). While SSMs achieve efficient parallelization through structured linear recurrences, this linearity constraint limits their expressive power and precludes modeling complex, nonlinear sequence-wise dependencies.
To address this, we present \method, a framework that breaks the sequence-parallelization barrier for nonlinear RNNs. Building on prior work, we cast the sequence of nonlinear recurrence relationships as a single system of equations, which we solve in parallel using Newton's iterations combined with custom parallel reductions. Our implementation achieves speedups of up to $665\times$ over na\"ive sequential application, allowing training nonlinear RNNs at unprecedented scales. To showcase this, we apply \method to adaptations of LSTM and GRU architectures, successfully training models of 7B parameters that attain perplexity comparable to similarly-sized Transformers and Mamba2 architectures.
To accelerate research in efficient sequence modeling, we release the \method codebase as an open-source framework for automatic training-parallelization of nonlinear RNNs, enabling researchers and practitioners to explore new nonlinear RNN models at scale.}

\metadata[Code]{\url{{https://github.com/apple/ml-pararnn}}}
\metadata[Correspondence]{\sffamily Federico Danieli: \url{f_danieli@apple.com}}
\date{\sffamily\today}

\begin{document}

\maketitle

\section{Introduction}
\label{sec:intro}

Since its introduction by \cite{attention}, the Transformer architecture has quickly imposed itself as the \emph{de-facto} choice for sequence modeling, surpassing previous state-of-the-art, RNN-based models such as GRU and LSTM \citep{gru,lstm}. One key reason behind the rapid adoption of Transformers lies in the efficiency of their application at training time: their core sequence mixer, the attention mechanism, can in fact be applied in parallel along the length of the input sequence. This effectively overcomes one main limitation of classical RNNs, whose application must be unrolled sequentially along the input sequence.

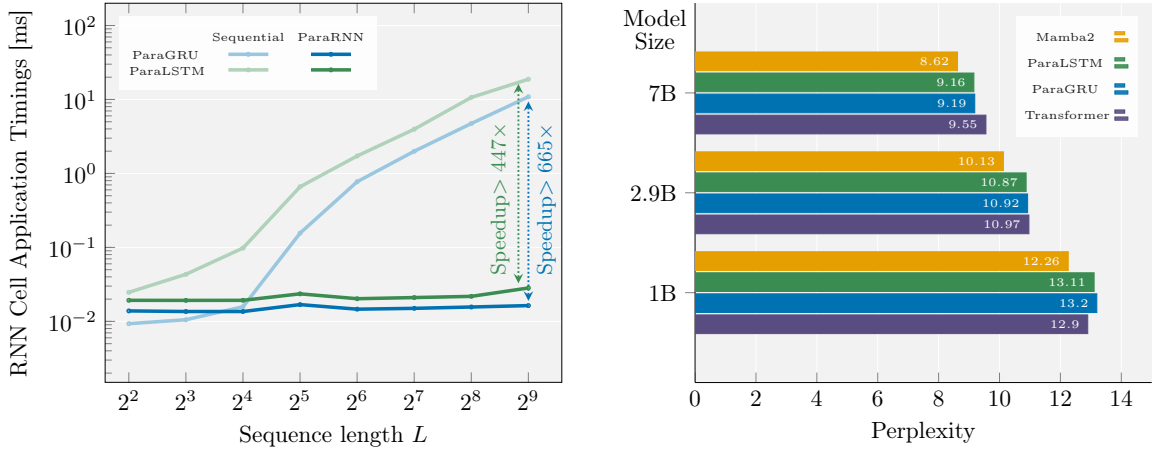
\begin{figure}[t!]
    \centering
    \resizebox{0.96\textwidth}{!}{\input{iclr2026/tikz_figs/fig_model_app_speedup}}
    \caption{Our \method framework makes it possible to apply classical RNNs in parallel, dramatically speeding up their training, and allowing them to be used competitively for language modeling.}
    \label{fig::speedup}
\end{figure}

In more recent times, however, interest in RNNs has been rekindled, largely due to their reduced memory footprint and improved efficiency at inference time. In particular, recent advancements in \emph{State Space Models} (SSMs) such as Mamba \citep{mamba,mamba2} have started gaining popularity and are imposing themselves as a potential alternative to Transformers, at least for small-to-mid-sized models \citep{zuo2024falconmambacompetitiveattentionfree}. To grant parallelization during training (and hence a comparable performance to attention), SSMs simplify the recurrence relationship at their core to one that is purely linear in the hidden state. This simplification enables leveraging associativity and parallel reduction operations to quickly compute the output of the application of an SSM to a whole input sequence in parallel along its length.
Despite the success of modern SSMs, the linearity constraint remains a limitation which hinders their expressive power \citep{illusionssm,cirone2025theoreticalfoundationsdeepselective}, and is dictated by necessity rather than choice. 

With \method, we aim to overcome the constraint of linearity for SSMs, while unlocking parallelization also for nonlinear RNNs, thus enriching the space of viable options for sequence modeling.

The method acts by re-casting the sequential application of an RNN to an input sequence of length $L$ as the solution of a nonlinear system of $L$ equations. Said solution is recovered via Newton's method: the system is linearized, and its solution approximated iteratively. In light of the Markovian nature of classical RNNs, the resulting linearized system obeys a specific block bi-diagonal structure, which we leverage to recover the solution efficiently, using custom high-performance implementations of parallel reduction operations. We provide a detailed description of the method in \cref{sec::method}.

For the sake of favoring the widespread adoption of this method among the research community, we make available the \method codebase. This takes care of automating the sequence-parallel application of an RNN cell, relying only on the definition of the RNN recurrence formula. The main structure of the framework is outlined in \cref{sec::software,app::code}.

Finally, we demonstrate the validity of \method in allowing the training of competitive RNNs for language modeling. To this end, in \cref{sec::LSTMD} we introduce two adaptations of the classical LSTM and GRU cells, tailored for efficient application within the \method framework. In \cref{sec::results}, we compare their performance against the Transformer and Mamba architectures, proving their competitiveness both in terms of runtime and of effectiveness in language modeling tasks.

\paragraph{Contributions} To summarize, our main contributions can be listed as follows:
\begin{enumerate}[leftmargin=.5cm]
	\item We adapt and develop efficient parallel reduction algorithms to enable training of nonlinear RNNs at unprecedented scales, expanding prior work on Newton-based RNN parallelization.
	\item We demonstrate for the first time that classical nonlinear RNN models can be trained on language modeling tasks at scales of 7B parameters, and achieve competitive performance with Transformers, thus enriching the space of available architecture choices for LLMs.
    \item We enable the exploration of new nonlinear RNN architectures for language modeling at scale, by introducing \method: a high-performance PyTorch+CUDA library that implements sequence-parallel training for any nonlinear cell from only the specification of its recurrence step, thereby automating the underlying parallelization complexity.
\end{enumerate}

\paragraph{Previous work}
There is a flourishing line of research around SSMs \citep{hippo,s4,s4d,lru}, which culminates more recently in Mamba and Mamba2 \citep{mamba,mamba2}, perhaps the most successful examples of modern SSMs. In a nutshell, SSMs revolve around a simplification of RNNs, that constrains the recurrence step to be purely \emph{linear} in the state. The associativity of linear operations enables then the use of \emph{parallel scan} \citep{parallelscanOG,parallelscan} to parallelize the RNN application. In principle, however, this approach to parallelization comes at the cost of expressivity: recent theoretical works warn in fact about the limitations of expressive power of linear SSMs \citep{illusionssm,huang2025understandinginputselectivitymamba,keiblinger2025recurrencecompleteframebasedactionmodels}. With our work, we effectively remove such linearity constraint.

Recent advantages in SSMs have also spawned a renewed interest in RNN adaptations. This is most noticeably represented by minGRU/minLSTM \citep{mingru}, and xLSTM \citep{xlstm}. However, the RNNs proposed in \cite{mingru} still preserve a linear recurrence at their core, although wrapped in additional nonlinearities that however---we remark---do \emph{not} interest its state. On the other hand, xLSTM proposes two alternative RNN cells: sLSTM and mLSTM. The first is purely linear, and parallelized similarly to Mamba; the second, being nonlinear, is still applied sequentially, without any attempt to parallelize its application sequence-wise: this comes with constraints on the size of the hidden states considered, otherwise its training-time application would become too costly.

The parallelization approach of \method takes inspiration from methods originally developed for time-parallelizing the solution of Ordinary Differential Equations (ODEs); see \cite{pint} for an overview. The specific idea of combining Newton's iterations and parallel reduction to solve for nonlinear sequential operations in parallel has been first introduced in \cite{deeppcr}, where it was applied to ResNets and Diffusion Models, but only hypothesized for language modeling and other sequence models. An application of this same approach has been investigated in \cite{parallelRNN} and improved in \cite{parallelRNN2}, where a GRU cell is trained in parallel for time-series modeling. There, the authors consider the use of a \emph{quasi}-Newton method, where a diagonal approximation of the full RNN Jacobian is employed for computational efficiency, but requires additional stabilization when directly applied to GRU. In this work, we show empirically that directly adapting the GRU and LSTM cells definition overcomes both these needs, while successfully extending to more complex cells that could not previously be tested.

Lastly, our work shares some looser connections with other approaches to speedup via parallelization of sequential operations. In \cite{song2021acceleratingfeedforwardcomputationparallel,Santilli2023}, Jacobi and Gauss-Seidel solvers are applied to accelerate ResNets applications, auto-regressive image generation, and Transformers inference; in \cite{shih2023parallelsamplingdiffusionmodels}, Picard iterations are used to parallelize denoising steps in Diffusion.

\section{Parallel application of nonlinear RNNs}
\label{sec::method}
In this work, we build upon the blueprint of \cite{deeppcr,parallelRNN}, where the sequential application of an RNN is re-cast as a system of nonlinear equations which is then solved via Newton iterations, and adapt it to LLM-scale training. For completeness, in the following we briefly describe the main components of the method used.

\paragraph{Newton's Method for Computing Nonlinear RNN Applications}
Consider the application of an RNN to an input sequence $[\bb{x}_l]_{l=1}^L$ of length $L$, whose elements are themselves vectors $\bb{x}_l\in\mathbb{R}^{d_{\text{in}}}$. The RNN cell tracks the evolution of a hidden state $\bb{h}_l\in\mathbb{R}^{d_{h}}$ through its recurrent application:
\begin{equation}
    \bb{h}_{l} = \bb{f}(\bb{h}_{l-1},\bb{x}_{l}), \qquad\forall l=1,\dots,L,
    \label{eqn::rnn_seq}
\end{equation}
starting from an all-zero initial state $\bb{h}_0=\bb{0}$, with $\bb{f}$ determining the cell action, or recurrence step. By collating the relationships in \cref{eqn::rnn_seq} we can define a single system of $L$ equations in the unknowns $[\bb{h}_l]_{l=1}^L$. This has the form
\begin{equation}
    \begin{cases}
        \bb{h}_1 - \bb{f}(\bb{0},\bb{x}_1) = \bb{0}\\
        \bb{h}_2 - \bb{f}(\bb{h}_1,\bb{x}_2)= \bb{0}\\
        \vdots\\
        \bb{h}_{L} - \bb{f}(\bb{h}_{L-1},\bb{x}_{L})= \bb{0}        
        \end{cases}
    \label{eqn::state}
\end{equation}
To solve this system, we rely on Newton's method \citep{nonlinearjacobi}. The method recovers an approximate solution of \cref{eqn::state} by iteratively solving its linearization, until convergence. More in detail, given the approximate solution at the $k$-th Newton iteration $[\bb{h}_l^k]_{l=1}^L$, this is updated solving
\begin{equation}
    \left[\begin{array}{cccc}
    I                                    &        &                                         &   \\
    -\left. \jacof{\bb{f}}\right|_{\bb{h}_{1}^k} & I      &                                         &   \\
                                         & \ddots & \ddots                                  &   \\
                                         &        & -\left. \jacof{\bb{f}}\right|_{\bb{h}_{L-1}^k} & I \\
    \end{array}\right]\left[\begin{array}{c}
        \delta\bb{h}_1^k\\
        \delta\bb{h}_2^k\\
        \vdots\\
        \delta\bb{h}_L^k\\
    \end{array}\right] = \left[\begin{array}{c}
        \bb{f}(\bb{0},\bb{x}_1) - \bb{h}_1^k\\
        \bb{f}(\bb{h}_1^k,\bb{x}_2) - \bb{h}_2^k\\
        \vdots\\
        \bb{f}(\bb{h}_{L-1}^k,\bb{x}_L) - \bb{h}_L^k\\
    \end{array}\right],
    \label{eqn::newton_it}
\end{equation}
so that $[\bb{h}_l^{k+1}]_{l=1}^L=[\bb{h}_l^k + \delta\bb{h}_l^k]_{l=1}^L$. In \cref{eqn::newton_it}, $\left.\jacof{\bb{f}}\right|_{\bb{h}_l^k}$ denotes the Jacobian of the recurrence step \cref{eqn::rnn_seq} with respect to the hidden state $\bb{h}$, evaluated at $\bb{h}_l^k$. Pseudo-code for the method is given in \cref{alg::newton}.

The second ingredient for effectively tackling the solution of \cref{eqn::state} consists of an efficient solver for the highly-structured, block bi-diagonal linear system in \cref{eqn::newton_it}. This is described in the following section.

\paragraph{Parallel Reduction for All-at-once Solution of Block Bi-diagonal Linear Systems}
As shown in the previous section, thanks to Newton's method, we can re-cast the application of a nonlinear RNN\footnote{While the method presented here considers the general case where the cell action \cref{eqn::rnn_seq} is \emph{nonlinear} in the state $\bb{h}_l$, it also naturally encompasses \emph{linear} RNNs, such as Mamba and other SSMs. In these cases, we have
$$\bb{f}_{\texttt{SSM}}(\bb{h}_{l-1},\bb{x}_l)=\bb{A}_l \bb{h}_{l-1} + \bb{B}_l \bb{x}_l,$$
and the Jacobians in \cref{eqn::newton_it} reduce to the state transition matrices themselves, $\left.\jacof{\bb{f}_{\texttt{SSM}}}\right|_{\bb{h}_{l-1}}\equiv \bb{A}_l$. Consequently, Newton's update reduces to the vanilla SSM application, and the target output is recovered in a single iteration.} into the iterative solution of a linear system in the form \cref{eqn::newton_it}. In principle, its solution $[\delta\bb{h}_l^k]_{l=1}^L$ could be explicitly recovered by forward substitution, sequentially unrolling the formula
\begin{equation}
    \delta\bb{h}_l^k =  \left. \jacof{\bb{f}}\right|_{\bb{h}_{l}^k}\delta\bb{h}_{l-1}^k +(\bb{f}(\bb{h}_{l-1}^k,\bb{x}_l) - \bb{h}_l^k),\qquad\forall l=1,\dots,L,
    \label{eqn::newton_it_seq}
\end{equation}
but this would defeat the purpose of parallelization. Instead, notice that \cref{eqn::newton_it_seq} itself corresponds to the application of a linear RNN, where the Jacobians $\left.\jacof{\bb{f}}\right|_{\bb{h}_{l}^k}$ cover the role of the state matrix, while the residuals $\bb{r}_l=\bb{f}(\bb{h}_{l-1}^k,\bb{x}_l) - \bb{h}_l^k$ cover that of the input (or forcing term). As such, the solution can be recovered explicitly by leveraging associativity of matrix multiplication when unrolling the recursion \cref{eqn::newton_it_seq}. We have in fact
\begin{equation}
\begin{split}
    \delta\bb{h}_l^k &=
    \left. \jacof{\bb{f}}\right|_{\bb{h}_{l}^k}\delta\bb{h}_{l-1}^k + \bb{r}_l
    =\left. \jacof{\bb{f}}\right|_{\bb{h}_{l}^k}\left(\left. \jacof{\bb{f}}\right|_{\bb{h}_{l-1}^k}\delta\bb{h}_{l-2}^k + \bb{r}_{l-1}\right) + \bb{r}_l
    =\ldots
    = \sum_{s=1}^l \prod_{r=0}^{l-s-1} \left.\jacof{\bb{f}}\right|_{\bb{h}_{l-r}^k} \bb{r}_s.
\end{split}
\label{eqn::newton_it_seq_unroll}
\end{equation}
There is some redundancy in the terms appearing in the sum in \cref{eqn::newton_it_seq_unroll} for different $l$'s: this is what allows specialized algorithms such as \emph{prefix sum} (also known as parallel reduction, or parallel scan \citep{parallelscanOG,parallelscan}) to compute the whole solution $[\delta\bb{h}_l^k]_{l=1}^L$ at once, in parallel, in $\mc{O}(\log_2L)$ steps, rather than sequentially in $\mc{O}(L)$ steps. Pseudo-code for a na\"ive implementation of the algorithm is provided in \cref{alg::pcr}, but we refer to \cref{app::code::solver} for the complete description of the more efficient CUDA implementation developed in this work.

\paragraph{Parallel Backward Pass Through RNN Applications}
Unlike the forward pass, which for general RNNs requires the solution of a nonlinear system, the backward pass through an RNN cell is an inherently linear operation. As such, we do not need to rely on Newton iterations, and the necessary gradients can be recovered directly in a single parallel reduction step. More in detail, given the partial derivatives of the loss function $\mc{L}$ with respect to the cell hidden states $[\partial_{\bb{h}_l}\mc{L}]_{l=1}^L$, computing the full gradients amounts to unrolling the backward recurrence (starting from $\nabla_{\bb{h}_{L}} \mathcal{L} = \partial_{\bb{h}_{L}} \mathcal{L}$)
\begin{equation}
\nabla_{\bb{h}_{l-1}} \mathcal{L} = \left. \jacof{\bb{f}}\right|_{\bb{h}_{l-1}}^\top \nabla_{\bb{h}_{l}} \mathcal{L} + \partial_{\bb{h}_{l-1}}\mc{L}, \qquad\forall l=L,\dots,1.
\label{eqn::rnn_bwd_seq}
\end{equation}
Notice this shares the same structure of \cref{eqn::newton_it_seq_unroll}, except it involves transposes of Jacobians, and it unrolls backwards. This notwithstanding, \cref{eqn::rnn_bwd_seq} can be solved using the same parallel reduction algorithm (\cref{alg::pcr}), with minimal modifications.

\begin{algorithmgroup}[t]
\begin{minipage}[t]{0.52\textwidth}
\begin{algorithm}[H]
    \renewcommand{\thealgocf}{\arabic{algocf}a}  
    \caption{Newton's method}
    \label{alg::newton}
    \addtocounter{algocf}{-1}  
    \SetKwFunction{ParallelSolve}{ParallelReduce}
    \tcc{Initialization:}
    Assemble initial guess $\bb{h}_{l}^0$\;
    \tcc{Newton iterations}
    \For{$k=1$ \KwTo $N_{\text{its}}$}{
        \tcc{Assemble system \cref{eqn::newton_it_seq}:}
        Compute Jacobians $\left. \jacof{\bb{f}}\right|_{\bb{h}_{l}^k}$\;
        Compute residuals $\bb{r}_l^k$\;
        \tcc{Solve in parallel}
        $\bb{h}_l^{k} \leftarrow \bb{h}_l^{k-1} + \ParallelSolve(\left. \jacof{\bb{f}}\right|_{\bb{h}_{l}^k},\bb{r}_l^k)$\;
    }
    \Return{$\bb{h}_l^{N_{\text{its}}}$}\;
\end{algorithm}
\end{minipage}
\hfill
\begin{minipage}[t]{0.42\textwidth}
\begin{algorithm}[H]
    \renewcommand{\thealgocf}{\arabic{algocf}b}  
    \caption{Parallel reduction}
    \label{alg::pcr}
    \addtocounter{algocf}{-1}  
    \KwIn{Jacobians $\left. \jacof{\bb{f}}\right|_{\bb{h}_{l}}$, residuals $\bb{r}_l$}
    \tcc{Initialization:}
    $A_l = \left. -\jacof{\bb{f}}\right|_{\bb{h}_{l}};\quad\delta\bb{h}_l = \bb{r}_l$\;
    \tcc{Reduction steps:}
    \For{$i=0$ \KwTo $\log_2L-1$}{
        \Parfor{$l=2^i$ to $L$}{
            $\delta\bb{h}_l \leftarrow \delta\bb{h}_l - A_l\delta\bb{h}_{l-2^i+1}$\;\label{alg::pcr::reduce_rhs}
            $A_l \leftarrow -A_lA_{l-2^i+1}$\;\label{alg::pcr::reduce_jac}
        }
    }
    \Return{$\delta\bb{h}_l$}\;
\end{algorithm}
\end{minipage}
\caption{Pseudo-code for the parallel application of a nonlinear RNN cell: we use Newton (left) as outer solver for the corresponding nonlinear system \cref{eqn::state}, while parallel reduction (right) is used as inner solver for the linearized system \cref{eqn::newton_it}.}
\label{alg::rnn_fwd}
\end{algorithmgroup}

\subsection{Limitations}
\label{sec::limitations}
While the \method framework described in this section is general enough to be applied to virtually any Markovian RNN, it presents two major limitations that must be kept into account when deploying it in practical applications.

\paragraph{Newton's Convergence} Our method relies on the convergence of Newton's iterations to recover the result from the RNN cell application. \citet{parallelRNN2} showed that, for systems like \cref{eqn::state}, convergence in $L$ Newton steps is guaranteed. However, such a slow convergence is not practical: it would break any advantage from parallelization, making a vanilla sequential application of the RNN cell more efficient. In practice, we need convergence in a small $\mc{O}(1)$ number of iterations. Fortunately, for the RNN considered in this work, \textit{three} iterations proved to be sufficient in all cases, but this must be verified for newly defined RNN wanting to leverage \method. We refer to \cref{app::newton} for additional information regarding the Newton solver setup and convergence in our application.

\paragraph{Computational Efficiency}
While \method allows to attain sequence-parallelizability, it does so at the cost of additional overhead computations, which would not be necessary in the sequential case. This is a common trade-off in the design of parallel algorithms. In our case, the brunt of this overhead is associated with the assembly and manipulation of $\jacof{\bb{f}}$ terms in \cref{eqn::newton_it_seq_unroll}. In fact, if we were to na\"ively consider RNNs with dense Jacobians, their memory requirement would scale as $\mc{O}(Ld_h^2)$, and each pair-wise multiplication (\cref{alg::pcr::reduce_jac} of \cref{alg::pcr}) would incur a computational cost of $\mc{O}(d_h^3)$, making it \emph{de-facto} unfeasible. In practice, then, one should inject some additional structure in the Jacobians definition, so to make these operations more tractable.
Notice that a similar problem arises when dealing with SSMs: to overcome this, e.g. in S4D \citep{s4d} the authors impose a diagonal structure on the SSM state matrix, to render the application of the prefix sum algorithm computationally feasible. This assumption is inherited by the Mamba model \citep{mamba}, for analogous considerations. Similarly, \citet{parallelRNN2} opt to use a \emph{quasi-}Newton method, where the RNN Jacobian is approximated with its diagonal, again rendering the resulting matrix multiplications leaner. In our case, we follow an approach similar to Mamba, as outlined next.

\section{Adapting GRU and LSTM for \method}
\label{sec::LSTMD}
As mentioned in \cref{sec::limitations}, the main obstacle to the efficient application of the prefix scan algorithm to \cref{eqn::newton_it_seq_unroll} lies in the computation of its product terms $\prod_{r=0}^{l-s-1} \left.\jacof{\bb{f}}\right|_{\bb{h}_{l-r}^k}$. Rather than considering Jacobians approximations as in \cite{parallelRNN2}, in our approach we choose instead to directly tune the RNN definition to yield Jacobians with a simplified structure. The adapted GRU and LSTM cells are described next.

\paragraph{RNN Cell Definitions}
In defining the main RNNs used in our work, we start from two of the most broadly established architectures available in the literature: the \emph{Gated Recurrent Unit} (GRU) \citep{gru} and the \emph{Long-Short Term Memory} (LSTM) \citep{lstm} cells. For the latter, we consider its variant equipped with peephole connections and combining input and forget gates \citep{lstm_cifg}. Their actions are defined as, respectively:

\begin{subequations}
\label{eqn::GRU_LSTM}
\begin{minipage}{0.4\textwidth}
\begin{equation}
\vcenter{\hbox{\rotatebox{90}{\footnotesize GRU}}}
\left\{\hspace{-0.22em}\begin{array}{r@{{}={}}l}
    \bb{z}_{l} & \sigma_g\left(A_z\bb{h}_{l-1} + B_z\bb{x}_l + \bb{b}_z\right)\\
    \bb{r}_{l} & \sigma_g\left(A_r\bb{h}_{l-1} + B_r\bb{x}_l + \bb{b}_r\right)\\
    \bb{c}_{l} & \sigma_h\left(A_c(\bb{h}_{l-1}\odot\bb{r}_{l}) + B_c\bb{x}_l + \bb{b}_c\right)\\    
    \multicolumn{2}{l}{\hspace{-0.39em}\colorbox{gray!30}{$\bb{h}_l = (\bb{1}-\bb{z}_l) \odot \bb{h}_{l-1} + \bb{z}_l \odot \bb{c}_l$}}
\end{array}\right.,
\label{eqn::GRU}
\end{equation}
\end{minipage}
\hfill
\begin{minipage}{0.56\textwidth}
\begin{equation}
\text{and}\qquad\quad
\vcenter{\hbox{\rotatebox{90}{\footnotesize LSTM}}}
\left\{\hspace{-0.22em}\begin{array}{r@{{}={}}l}
    \bb{f}_{l} & \sigma_g\left(A_f\bb{h}_{l-1} + B_f\bb{x}_l + C_f \bb{c}_{l-1} + \bb{b}_f\right)\\
    \bb{z}_{l} & \sigma_z\left(A_z\bb{h}_{l-1} + B_z\bb{x}_l + \bb{b}_z\right)\\
    \multicolumn{2}{l}{\hspace{-0.238em}\colorbox{gray!30}{$\bb{c}_l = \bb{f}_l \odot \bb{c}_{l-1} + (\bb{1}-\bb{f}_l) \odot \bb{z}_l$}}\\
    \bb{o}_{l} & \sigma_g\left(A_o\bb{h}_{l-1} + B_o\bb{x}_l + C_o \bb{c}_{l} + \bb{b}_o\right)\\
    \multicolumn{2}{l}{\hspace{-0.25em}\colorbox{gray!30}{$\bb{h}_l = \bb{o}_l \odot \sigma_h(\bb{c}_l)$}}
\end{array}\right.
\label{eqn::LSTM}
\end{equation}
\end{minipage}
\end{subequations}

for $l=1\dots L$. The highlighted equations represent the actual hidden state of the two cells (that is, the variables involved in the recurrence step), while the other ones are auxiliary temporary variables needed to perform the update itself. Notice in particular that for LSTM the hidden state is given by the collation of two variables: $\hat{\bb{h}}_l=[\bb{c}_l^\trans,\bb{h}_l^\trans]^\trans\in\mathbb{R}^{2d_h}$.



\paragraph{Cell Jacobians}
To apply the \method machinery described in \cref{sec::method} to the cells in \cref{eqn::GRU_LSTM}, we need to compute the Jacobians of their recurrence step with respect to the hidden state. These can be recovered explicitly with some algebraic operations, and come in the form, respectively:

\begin{subequations}
\label{eqn::GRU_LSTM_jac}
\begin{align}
\jacof{\text{GRU}}  =&
\diag(\bb{1}-\bb{z}_l) + \diag((\bb{c}_l-\bb{h}_{l-1})\odot\sigma_g'(\hat{\bb{z}}_l))A_z\notag\\
&+ \diag(\bb{z}_l\odot\sigma_h'(\hat{\bb{c}}_l))A_c(\diag(\bb{r}_l)+\diag(\bb{h}_{l-1}\odot\sigma_g'(\hat{\bb{r}}_l))A_r),\label{eqn::GRU_jac}\\
\text{and}\qquad\quad\jacof{\text{LSTM}} =&
\left[\begin{array}{c|c}
\jacof{\bb{cc}} & \jacof{\bb{ch}}\\\hline
\jacof{\bb{hc}} & \jacof{\bb{hh}}
\end{array}\right],\quad\text{with}\quad
\left\{\begin{array}{r@{{}{}}l}
\jacof{\bb{cc}} =&\diag(\bb{f}_l) + \diag((\bb{c}_{l-1}-\bb{z}_l)\odot\sigma_g'(\hat{\bb{f}}_l))C_f\\
\jacof{\bb{ch}} =&\diag((\bb{c}_{l-1}-\bb{z}_l)\odot\sigma_g'((\hat{\bb{f}}_l)))A_f \\&+ \diag((\bb{1}-\bb{f}_l)\odot\sigma_z'(\hat{\bb{z}}_l))A_z\\
\jacof{\bb{hc}} =&(\diag(\sigma_h(\bb{c}_l)\odot\sigma_g'(\hat{\bb{o}}_l))C_o \\&+ \diag(\bb{o}_l\odot\sigma_h'(\bb{c}_l)))\jacof{\bb{cc}}\\
\jacof{\bb{hh}} =&\diag(\sigma_h(\bb{c}_l)\odot\sigma_g'(\hat{\bb{o}}_l))(A_o+C_o\jacof{\bb{ch}}) \\&+ \diag(\bb{o}_l\odot\sigma_h'(\bb{c}_l))\jacof{\bb{ch}}
\end{array}\right., \label{eqn::LSTM_jac}
\end{align}
\end{subequations}

where $\diag(\bb{v})$ denotes the diagonal matrix with the elements from $\bb{v}$ as its diagonal, $\sigma_{*}'$ is the derivative of $\sigma_{*}$, and for a given temporary variable $\bb{v}$, $\hat{\bb{v}}$ denotes its value before the nonlinearity application. Notice that the LSTM Jacobian features a $2\times2$-block structure, directly stemming from the fact that the hidden state is split into two variables---so that each sub-Jacobian $J_{*\star}$ indicates the partial derivative $\partial(*)_{l}/\partial(\star)_{l-1}$.

\paragraph{Jacobians Structure Simplification}
Our goal is to make Jacobians in \cref{eqn::GRU_LSTM_jac} leaner to store and to multiply together. Notice that the overall structure of these Jacobians is ultimately linked to that of the various state and peephole-connection matrices $A_{*}, C_{\star}$, since all other matrices involved are purely diagonal. An apt simplification consists then in picking these matrices, too, to be diagonal; in other words, we substitute in \cref{eqn::GRU_LSTM}:
\begin{equation}
    A_* = \diag(\bb{a}_*), \quad C_{\star} = \diag(\bb{c}_{\star}), \qquad \text{for some learnable parameters}\qquad \bb{a}_*,\bb{c}_{\star}\in\mathbb{R}^{d_h}.
    \label{eqn::GRU_LSTM_simp}
\end{equation}
This reduces $\jacof{\text{GRU}}$ in \cref{eqn::GRU_LSTM_jac} to a diagonal matrix, and further simplifies the $2\times2$-block structure of $\jacof{\text{LSTM}}$ to one with diagonal blocks. In both cases, the Jacobians thus simplified occupy only $\mc{O}(d_h)$ memory, and can be multiplied together with $\mc{O}(d_h)$ complexity, in a perfectly parallelizable fashion over the hidden state components.

While this simplification allows us to satisfy the requirements (outlined in \cref{sec::limitations}) to apply parallel scan effectively, we point out that it has the effect of inhibiting any mixing of the hidden state components \emph{within} the RNN cell: indeed, with \cref{eqn::GRU_LSTM_simp} we are effectively reducing a single $d_h$-dimensional RNN cell to $d_h$ independent $1$-dimensional ones, potentially affecting expressivity. In practice, however, feature mixing is taken care by subsequent MLP layers. Moreover, this simplification is in line with practices for modern SSM, which analogously consider diagonal state matrices.
We also point out that an LSTM with a similarly simplified structure was considered by \citet{sru}, except there the goal was to improve efficiency in the \emph{sequential} application of the LSTM cell, while in our work the focus is to allow a computationally-efficient parallel reduction operation to \emph{parallelize} the cell application along the sequence length. More aligned with our goal is the recent work in \cite{farsang2025scalingliquidresistanceliquidcapacitancenetworks}, where the authors similarly consider a diagonalized RNN cell to simplify the Jacobian structure in the Newton routine to aid parallelizability.


\section{Efficient implementation of parallel reduction}
\label{sec::software}
As part of the main contributions of this work, we make available \method, a fully modular, PyTorch+CUDA-based software framework allowing to readily parallelize the application of virtually \emph{any} user-defined RNN cell. The user needs only to prescribe the recurrence step defining the cell action (including any parameters necessary to its definition), and the \method framework takes care of automatically assembling the resulting all-at-once system. Particularly, it leverages \texttt{autograd} to assemble the Jacobians required in \cref{eqn::newton_it}, and then proceeds to efficiently solve the system via Newton's method and parallel reduction (\cref{alg::rnn_fwd}), parallelizing the RNN cell application.

In \method we provide three different implementations of the parallel solver, allowing users to trade off ease-of-use against performance:
\begin{enumerate}[leftmargin=.5cm]
    \item \textbf{Pure PyTorch} A reference implementation of parallel reduction which only relies on native PyTorch operations, and is mainly thought for prototyping, debugging, and exploring new RNN cell definitions. While not directly optimized for performance, this version provides full automatic differentiation support and seamless integration with existing PyTorch models. By manually providing the explicit formula for the Jacobians assembly, the user can still choose to boost performance by overriding the default \texttt{autograd}-based ones.\label{imp::torch}
    \item \textbf{CUDA-Accelerated Parallel Reduction} A performant implementation featuring a custom CUDA kernel for the parallel reduction solver, specialized for Jacobians with diagonal or $N\times N$-block-diagonal structure as in \cref{eqn::GRU_LSTM_jac}. Jacobian assembly remains in PyTorch, preserving automatic differentiation capabilities while accelerating the computational bottleneck. The CUDA solver employs a hardware-aware, hybrid algorithm combining forward substitution and parallel reduction at distinct levels in the GPU hierarchy (\emph{thread}, \emph{warp}, \emph{block} and \emph{device}, see \cite{cuda2025}). The goal is to maximize register utilization and take full advantage of efficient warp-level directories, while minimizing shared and global memory traffic: we refer to \cref{app::code::solver} for full implementation details.\label{imp::CUDA}
    \item \textbf{Fully-fused} Our most performant implementation where the whole Newton routine, including iterative Jacobian assembly and parallel solution, is fused within a single CUDA kernel. This eliminates intermediate memory traffic and overhead from kernel function calls, although it requires users to provide a CUDA implementation of the RNN cell. Nonetheless, we still preserve a modular approach, so that the user only needs to define those operations which directly pertain to the cell itself (namely, recurrence step and Jacobian assembly: see \cref{app::code::example}). Our code provides a reference implementation for the \GRUD and \LSTMD cells described in \cref{sec::LSTMD}.\label{imp::CUDA-fused}
\end{enumerate}

\section{Results}
\label{sec::results}
The experiments reported in this section aim at evaluating the competitiveness of \method in Language Modeling tasks, particularly when applied to the \GRUD and \LSTMD cells. To this purpose, we focus on two main metrics. First, in \cref{sec::efficiency}, we investigate the computational cost (primarily in terms of wall-time) associated with utilizing \method: our goal is to demonstrate that, by unlocking sequence-parallelizability, \method allows to (i) train nonlinear-RNN-based models with a runtime comparable to that of Transformer and Mamba architectures; and (ii) achieve inference speed much faster than Transformer, and similar to Mamba. Secondly, in \cref{sec::performance} we analyze the overall performance of \GRUD and \LSTMD as language models, and show that they can achieve competitive perplexity and downstream task performance.

\subsection{Implementation performance}
\label{sec::efficiency}
\begin{figure}[b!]
    \centering
    \resizebox{0.96\textwidth}{!}{\input{iclr2026/tikz_figs/fig_kernel_time_and_model_app_vs_seqlength}}
    \caption{Timing results for the parallel reduction operations \cref{alg::pcr} (left), and the full parallel RNN application \cref{alg::newton} (right), applied to sequences of varying length. Colors refer to different variants of the solvers: for diagonal Jacobians (blue, used by \GRUD) and 2x2 block-diagonal ones (green, used by \LSTMD). Heavier lines refer to progressively more efficient implementations: in PyTorch, in CUDA, and (for the RNN forward pass) with the fully-fused CUDA kernel. For reference, equivalent timing results for Mamba are also included (yellow). Actual speedup measurements reported in the main text refer to $L=2^9$ (highlighted).}
    \label{fig::kernel_and_model_own_vs_mamba}
\end{figure}
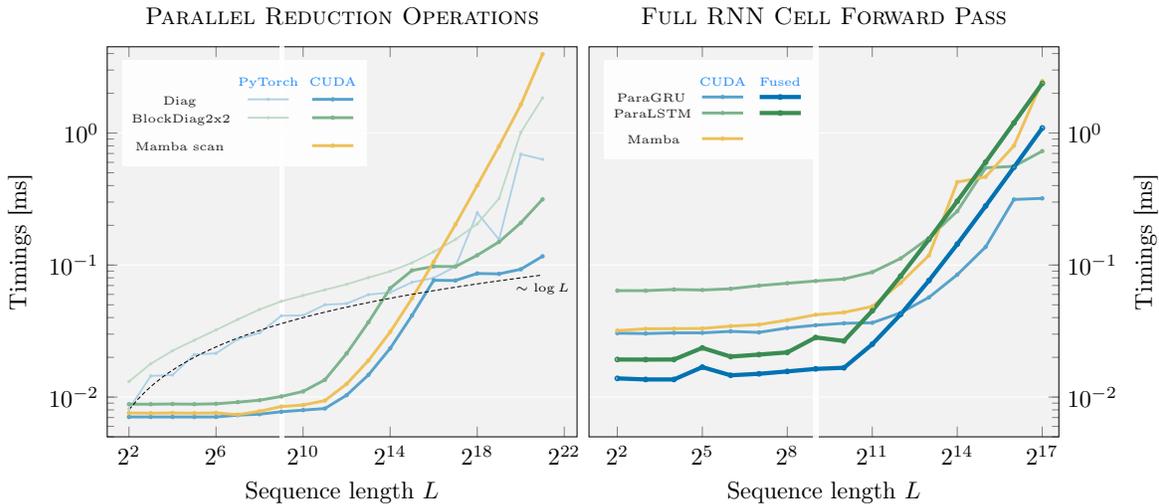

\paragraph{Competitive Runtime}
Here we report profiling results for the main functions involved in the application of \method. First, in \cref{fig::kernel_and_model_own_vs_mamba} (left) we time our different implementations of the parallel reduction operation \cref{alg::pcr}. Notably, our CUDA implementation for diagonal Jacobians slightly improves on the equivalent \texttt{parallel\_scan} shipped with Mamba ($\sim1.1\times$ speedup at $L=2^9$); the one for block-diagonal Jacobians is instead slower ($\sim0.84\times$ slowdown at $L=2^9$)---as expected, being it more memory and computationally intensive. Either case, the brunt of the kernel cost is associated with the memory reading overhead, rather than the computation itself, as indicated by the lines being mostly flat until $L\sim2^{11}$. The inherent complexity of the operation is instead more visible from the profiling of the PyTorch implementation: this closely follows a logarithmic growth, before collapsing to linear (from $L\gtrsim2^{18}$), when the GPU is close to capacity and operations must be sequentialized.
The profiling curves also mirror the three regimes of our parallel scan implementation described in \cref{app::code::solver}: (i) the sequence fits in a single GPU block and is analyzed in parallel within it ($L\lesssim2^{10}$); (ii) the sequence is analyzed in parallel within each block, but sequentially over a few blocks ($2^{10}\lesssim L\lesssim2^{16}$, where the curve rises to a linear regime); and (iii) the sequence is analyzed in parallel across blocks ($L\gtrsim2^{16}$, where we observe the curve flattening again to a $\log$). By contrast, Mamba's implementation collapses to sequential across blocks for $L\gtrsim2^{10}$.

Similar observations on relative performance hold also when we consider the complete parallel application of the RNN cell forward pass---that is, the full Newton routine \cref{alg::rnn_fwd} instead of the sole parallel reduction operation \cref{alg::pcr}. Profiling results for this setting are reported in \cref{fig::kernel_and_model_own_vs_mamba} (right). Already our implementation combining Jacobians assembly in PyTorch and parallel reduction in CUDA provides runtimes equivalent to the Mamba SSM application: this is basically comparable to applying \GRUD (we attain a $\sim1.2\times$ speedup for $L=2^9$), while \LSTMD still lags behind (with a $\sim0.5\times$ slowdown), again following expectations. However, it is by fully-fusing the whole Newton routine in CUDA that we achieve the largest speedups with respect to Mamba: of $\sim2.6\times$ and $\sim1.5\times$ for \GRUD and \LSTMD respectively, at $L=2^{9}$.

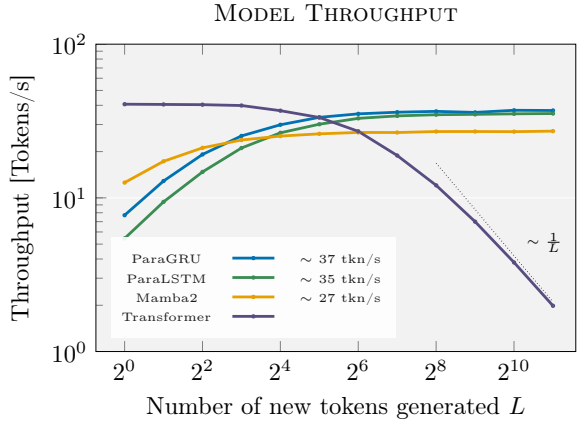
\begin{wrapfigure}[16]{r}{0.5\textwidth}
    \centering
    \resizebox{0.48\textwidth}{!}{\input{iclr2026/tikz_figs/fig_model_throughput}}
    \caption{Tokens throughput at generation for \GRUD (blue), \LSTMD (green), Mamba (yellow), and Transformers (purple).}
    \label{fig::throughput_own_vs_mamba}
\end{wrapfigure}

\paragraph{Fast Inference} For inference, we observe a runtime similar to that of Mamba, as shown in \cref{fig::throughput_own_vs_mamba}: at regime, we attain a throughput of $\sim35-37$ tokens per second, versus the $\sim27$ of Mamba. Notice moreover that, unlike for Transformers, for RNNs the time necessary to autoregressively produce the next token in the output \emph{does not grow with sequence length}.

Overall, the measurements reported in this section support our claim that \method represents a viable option for accelerating the application of RNNs, with runtime competitive with (and often beating) Mamba. We refer to \cref{app::profiling} for additional timing results and more details on the setup for the experiments considered here.

\subsection{Language modeling performance}
\label{sec::performance}
Having established runtime parity between Mamba and our RNN cells, we now analyze a different axis of \GRUD and \LSTMD effectiveness: their performance as actual language models. Our goal is to show that, with sequentiality constraints removed and training at scale enabled, even classical RNNs can achieve competitive performance in language modeling tasks.

\paragraph{The Importance of Nonlinearities}
We begin by evaluating \GRUD and \LSTMD on synthetic tasks that isolate fundamental capabilities required for language modeling. We focus on retrieval and state-tracking tasks (MQAR, k-hop, Parity, \citep{arora2023zoology})---where linear SSMs like Mamba are known to struggle~\citep{illusionssm}---to establish the critical motivation behind our work: \emph{nonlinear} RNNs provide computational expressivity that \emph{linear} RNNs fundamentally lack, and hence merit continued development.

\begin{wraptable}[9]{r}{0.38\textwidth}
\footnotesize
\vskip -0.18in
\centering
\caption{Single-layer accuracy on synthetic tasks. $()^\dagger$ denotes results computed training only the RNN cell (see \cref{app::synthetic} for details).}
\label{tab::synthetic}
\begin{tabular}{l *{3}{c@{\hspace{4pt}}}}
    \toprule
    Model & MQAR & $k$-hop  & Parity$^\dagger$\\
    \midrule
    Transformer & 100\% &  78\% & 53\%\\
    Mamba2      & 100\% &  98\% & 51\%\\ 
    \GRUD       & 100\% & 100\% & 100\%\\
    \LSTMD      & 100\% & 100\% & 100\%\\
    \bottomrule
\end{tabular}
\end{wraptable}

To directly assess the expressivity of the sole recurrent cell, we use single-layer architectures consisting only of an embedding layer, normalization, the RNN cell considered (or Attention, for comparison), and a linear layer---see a detailed description in \cref{app::synthetic}. As shown in \cref{tab::synthetic}, \GRUD and \LSTMD can solve all three tasks perfectly, while Mamba2---being purely linear---fails on Parity and (similarly to Transformers), does not reach a perfect solution on $k$-hop. Overall, the results highlight the key role of nonlinearities in boosting expressivity.

\paragraph{Matching Transformers Without Attention} Moving beyond synthetic tasks, we now evaluate our cells on language modeling to assess their effectiveness. To this end, we train four models: our \GRUD and \LSTMD presented in \cref{sec::LSTMD}, plus a Transformer with the DCLM architecture~\citep{li2024datacomplm} and Mamba2~\citep{mamba2} as baselines. For each architecture, we consider four different model sizes (of 400M, 1B, 2.9B, and 7B parameters), and follow Chinchilla-optimal~\citep{hoffmann2022trainingcomputeoptimallargelanguage} scaling for their training setup, using SlimPajama~\citep{cerebras2023slimpajama} without the Books3 split as training dataset: a complete description of the training procedure, including hyperparameters configuration and training times is reported in \cref{app::llm}. In designing our RNN models, we build upon the DCLM Transformer backbone, replacing attention with our RNN cells, but including also the causal convolution and gated residual layers from Mamba: detailed architecture specifications are listed in \cref{app::llm}, and a schematic of the RNN block is illustrated in \cref{fig::RNN_model_schematics}.

\begin{table*}[b!]
\scriptsize 
\setlength{\tabcolsep}{6pt}
\centering
\caption{Perplexity, parameters count, and evaluation scores on \texttt{lm-eval-harness} \citep{eval-harness} tasks, for 7B models. Accuracies in percentages. Shot counts are reported in brackets.}
\label{tab::lm-eval-harness}
\begin{tabular}{l cc c@{\hspace{4pt}}c c@{\hspace{4pt}}c c@{\hspace{4pt}}c c@{\hspace{4pt}}c cc }
    \toprule
    & & & \multicolumn{2}{c}{$\uparrow$~Arc-C} & \multicolumn{2}{c}{$\uparrow$~HSwag} & \multicolumn{2}{c}{$\uparrow$~OBQA} & \multicolumn{2}{c}{$\uparrow$~WinoG} & $\uparrow$~PiQA & $\uparrow$~MMLU \\
    Model & \#params & $\downarrow$~PPL & (25) & (3) & (10) & (0) & (10) & (0) & (5) & (0) &  (0) &  (0) \\
    \midrule
    Mamba2      & 6.96B & 8.62 & 40.02 & 39.59 & 69.78 & 69.68 & 42.20 & 42.20 & 65.19 & 63.77 & 76.66  & 26.61 \\ 
    \LSTMD      & 6.76B & 9.16 & 37.46 & 36.52 & 62.47 & 62.85 & 42.20 & 38.80 & 57.70 & 59.12 & 75.19  & 25.31 \\
    \GRUD       & 6.76B & 9.19 & 39.68 & 36.77 & 65.85 & 65.75 & 42.20 & 40.40 & 61.40 & 59.83 & 76.66  & 25.29 \\
    Transformer & 6.89B & 9.55 & 34.30 & 33.36 & 62.98 & 62.20 & 40.00 & 37.20 & 61.48 & 60.85 & 74.97  & 23.12 \\
    \bottomrule
\end{tabular}
\end{table*}

\Cref{tab::lm-eval-harness} presents the downstream task performance of our 7B-parameters trained models on reference tasks from the \texttt{lm-eval-harness} suite \citep{eval-harness} (see \cref{tab::lm-eval-harness_complete} for complete results across scales). Results reveal a consistent ordering, in line with the perplexities reported in \cref{fig::speedup}: Mamba2 achieves the strongest performance, followed by our \GRUD and \LSTMD models, with the DCLM Transformer trailing behind. To properly contextualize these results, we remark that our goal is not to propose a novel, more performant RNN, but to prove that also classical nonlinear RNNs can be trained efficiently at scale through parallelization, achieving performance competitive with modern language models. Ultimately, we are showing that the training inefficiencies that historically sidelined classic RNNs can be overcome, reviving them as viable options for language modeling.


\section{Conclusion and future work}
In this work, we demonstrate that the training-parallelization barrier for nonlinear RNNs can be effectively overcome even at scale, the linearity constraint no longer precluding additional explorations in efficient sequence modeling. Our empirical results show that even classical RNNs, when trained with \method, can achieve performance competitive with modern architectures---crucially, without sacrificing training efficiency.
This parity represents an important first step: it establishes that nonlinear RNNs are again a viable alternative in the modern landscape of sequence modeling.

In addition, the \method codebase makes this alternative readily available to practitioners, while delivering performance that matches or exceeds existing parallel implementations. By proving that classical RNNs can compete on equal footing with current architectures once computational barriers are removed, we aim for this work to further reignite exploration into nonlinear recurrent models.

\clearpage


\section*{Acknowledgments}
We would like to extend our thanks to: Hadi Pour Ansari, for all the help during the reviewing process of this manuscript; Eeshan Gunesh Dhekane and Jagrit Digani, for the extremely valuable feedback, particularly regarding the CUDA kernels implementation; and Nick Apostoloff and Jerremy Holland, for their support throughout this project.

\bibliography{iclr2026/iclr2026_conference}
\bibliographystyle{iclr2026/iclr2026_conference}

\newpage
\appendix
\input{apple_template/appendix}

\end{document}

%% file: apple_template/apple_preamble.tex
\usepackage{lipsum}
\usepackage{verbatim}

\usepackage{hyphenat}
\usepackage{amssymb, amsmath, amsfonts, amsopn, amsthm}
\usepackage{array, cases, stmaryrd}
\usepackage{mathtools, xfrac, scalerel}

\usepackage{xcolor}
\usepackage[most]{tcolorbox}
\input{iclr2026/applecolors}
\newtcolorbox{summarybox}
{
colback=AppleBlue!5,
colframe=AppleBlue5!90!black,
fonttitle=\bfseries,
title=Summary}
\colorlet{gruColor}{DeepBlue}
\colorlet{lstmColor}{AppleGreen5}
\colorlet{mambaColor}{DeepOrange}
\colorlet{transformerColor}{ApplePurple6}

\usepackage[linesnumbered, ruled, resetcount, vlined]{algorithm2e}
\SetKwFor{Parfor}{parfor}{do}{}

\SetCommentSty{mycodecomment}
\SetKwComment{tcc}{// }{}
\usepackage[frozencache,cachedir=.]{minted}
\definecolor{codebackground}{gray}{0.92}
\setminted{bgcolor=codebackground, style=friendly, fontsize=\scriptsize}

\usepackage{booktabs,colortbl,makecell,multirow}

\usepackage{graphicx}
\usepackage{wrapfig}
\usepackage{tikz,tikz-3dplot}
\usepackage{pgfplots,pgfplotstable}
\usepgfplotslibrary{patchplots,colorbrewer,fillbetween,units,groupplots}
\pgfplotsset{compat=1.8}
\usepackage{arydshln,subcaption}
\usetikzlibrary{math, matrix, fadings, shapes, calc, fit, backgrounds, arrows,arrows.meta, decorations.pathreplacing, decorations.pathmorphing, decorations.markings, positioning}
\usepackage{listofitems,siunitx,enumerate,tabu}
\pgfdeclarelayer{background}
\pgfdeclarelayer{foreground}
\pgfsetlayers{background,main,foreground}   
\usetikzlibrary{external}
\pgfkeys{
    /tikz/external/prefix=figures/,
    /tikz/external/mode=list and make
}

\usepackage{newtxtt}
\usepackage{diagbox}
\usepackage{tabularx}
\usepackage{silence}
\usepackage{dsfont}
\usepackage{xfakebold}

\definecolor{textgray}{HTML}{6E6E73}
\makeatletter
\patchcmd{\wrong@fontshape}{\@gobbletwo}{}{}{}
\makeatother
\numberwithin{equation}{section}
\makeatletter
\AtBeginDocument{
  \urlstyle{sf}
  
}
\makeatother

\definecolor{light}{RGB}{125, 125, 125}

\newtcolorbox[auto counter]{pbox}[2][]{
  colback=white,
  title=Code~\thetcbcounter: #2,
  #1,fonttitle=\sffamily,
  fontupper=\sffamily,
  arc=2pt,
  colframe=bgcolor,
  coltitle=fgcolor,
  colbacktitle=bgcolor,
  toptitle=0.25cm,
  bottomtitle=0.125cm
}

\makeatletter
\newcommand\applefootnote[1]{%
  \begingroup
  \renewcommand\thefootnote{}%
  \renewcommand\@makefntext[1]{\noindent##1}%
  \footnote{#1}%
  \addtocounter{footnote}{-1}%
  \endgroup
}
\makeatother

\definecolor{cverbbg}{gray}{0.90}

\DeclareMathAlphabet\mathbfcal{OMS}{cmsy}{b}{n}

\DeclareMathOperator*{\diag}{\mathrm{diag}}

\newcommand{\bb}[1]{{\boldsymbol{#1}}}
\newcommand{\mc}[1]{{\mathcal{#1}}}

\usepackage{url}
\usepackage{hyperref}
\usepackage{cleveref}
\usepackage{enumitem}

\crefformat{equation}{(#2#1#3)} 

\crefname{appendix}{App.}{App.}
\Crefname{appendix}{Appendix}{Appendices}
\crefname{section}{Sec.}{Sec.}
\Crefname{section}{Section}{Sections}
\crefname{table}{Tab.}{Tab.}
\Crefname{table}{Table}{Tables}
\crefname{figure}{Fig.}{Fig.}
\Crefname{figure}{Figure}{Figures}
\crefname{algorithm}{Alg.}{Alg.}
\Crefname{algorithm}{Algorithm}{Algorithms}
\crefname{theorem}{Thm.}{Thm.}
\Crefname{theorem}{Theorem}{Theorems}
\crefname{remark}{Rmk.}{Rmk.}
\Crefname{remark}{Remark}{Remarks}
\crefname{assumption}{assumption}{assumption}
\Crefname{assumption}{Assumption}{Assumptions}
\crefname{algocf}{Alg.}{Alg.}
\Crefname{algocf}{Algorithm}{Algorithms}
\crefname{AlgoLine}{line}{lines}
\Crefname{AlgoLine}{Line}{Lines}

\newcounter{myalg}
\AtBeginEnvironment{algorithmic}{\refstepcounter{myalg}}
\makeatletter
\@addtoreset{ALC@unique}{myalg}
\makeatother
\makeatletter
\newenvironment{algorithmgroup}[1][t]{%
  \@float{algocf}[#1]
}{%
  \end@float%
}
\makeatother

\newcommand{\jacof}[1]{J_{#1}}
\newcommand{\trans}{\top}

\usepackage{xspace}
\newcommand{\method}{{ParaRNN}\xspace}
\newcommand{\LSTMD}{{ParaLSTM}\xspace}
\newcommand{\GRUD}{{ParaGRU}\xspace}

\usepackage{placeins}
\usepackage[bottom]{footmisc}

%% file: iclr2026/applecolors.tex
\usepackage{xcolor}
\definecolor{AppleWhite}{RGB}{255,255,255}
\definecolor{ApplePrimaryCoolGray}{RGB}{116,128,139}
\definecolor{AppleCoolGray1}{RGB}{199,209,214}
\definecolor{AppleCoolGray2}{RGB}{147,174,190}
\definecolor{AppleCoolGray3}{RGB}{124,147,160}
\definecolor{AppleCoolGray4}{RGB}{92,102,109}
\definecolor{AppleCoolGray5}{RGB}{78,93,100}
\definecolor{AppleCoolGray6}{RGB}{53,60,65}
\definecolor{AppleBlack}{RGB}{0,0,0}
\definecolor{AppleSecondaryChartGray}{RGB}{168,168,168}
\definecolor{AppleChartGray2}{RGB}{233,233,233}
\definecolor{AppleChartGray3}{RGB}{211,211,211}
\definecolor{AppleChartGray4}{RGB}{190,190,190}
\definecolor{AppleChartGray5}{RGB}{140,140,140}
\definecolor{AppleChartGray6}{RGB}{102,102,102}
\definecolor{AppleChartGray7}{RGB}{64,64,64}
\definecolor{ApplePrimaryChartBlue}{RGB}{84,151,193}
\definecolor{AppleBlue2}{RGB}{212,229,239}
\definecolor{AppleBlue3}{RGB}{169,202,223}
\definecolor{AppleBlue4}{RGB}{127,177,209}
\definecolor{AppleBlue5}{RGB}{71,130,166}
\definecolor{AppleBlue6}{RGB}{55,99,128}
\definecolor{AppleBlue7}{RGB}{45,72,89}
\definecolor{ApplePrimaryChartGreen}{RGB}{83,172,121}
\definecolor{AppleGreen2}{RGB}{212,234,221}
\definecolor{AppleGreen3}{RGB}{169,213,188}
\definecolor{AppleGreen4}{RGB}{126,193,155}
\definecolor{AppleGreen5}{RGB}{58,140,82}
\definecolor{AppleGreen6}{RGB}{39,102,54}
\definecolor{AppleGreen7}{RGB}{29,58,31}
\definecolor{ApplePrimaryChartYellow}{RGB}{253,195,93}
\definecolor{AppleYellow2}{RGB}{254,240,214}
\definecolor{AppleYellow3}{RGB}{254,224,174}
\definecolor{AppleYellow4}{RGB}{254,210,134}
\definecolor{AppleYellow5}{RGB}{230,168,69}
\definecolor{AppleYellow6}{RGB}{191,131,46}
\definecolor{AppleYellow7}{RGB}{153,107,54}
\definecolor{ApplePrimaryChartOrange}{RGB}{250,151,92}
\definecolor{AppleOrange2}{RGB}{254,229,214}
\definecolor{AppleOrange3}{RGB}{252,203,173}
\definecolor{AppleOrange4}{RGB}{252,178,133}
\definecolor{AppleOrange5}{RGB}{227,121,68}
\definecolor{AppleOrange6}{RGB}{191,87,46}
\definecolor{AppleOrange7}{RGB}{143,59,36}
\definecolor{ApplePrimaryChartRed}{RGB}{227,94,105}
\definecolor{AppleRed2}{RGB}{248,215,217}
\definecolor{AppleRed3}{RGB}{241,174,180}
\definecolor{AppleRed4}{RGB}{234,135,143}
\definecolor{AppleRed5}{RGB}{196,63,77}
\definecolor{AppleRed6}{RGB}{153,35,53}
\definecolor{AppleRed7}{RGB}{102,19,43}
\definecolor{ApplePrimaryChartPurple}{RGB}{161,150,204}
\definecolor{ApplePurple2}{RGB}{231,228,242}
\definecolor{ApplePurple3}{RGB}{208,202,229}
\definecolor{ApplePurple4}{RGB}{185,176,217}
\definecolor{ApplePurple5}{RGB}{128,113,171}
\definecolor{ApplePurple6}{RGB}{89,76,128}
\definecolor{ApplePurple7}{RGB}{62,46,101}
\definecolor{AppleCoolGray}{RGB}{116,128,139}
\definecolor{AppleChartGray}{RGB}{168,168,168}
\definecolor{AppleBlue}{RGB}{84,151,193}
\definecolor{AppleGreen}{RGB}{83,172,121}
\definecolor{AppleYellow}{RGB}{253,195,93}
\definecolor{AppleOrange}{RGB}{250,151,92}
\definecolor{AppleRed}{RGB}{227,94,105}
\definecolor{ApplePurple}{RGB}{161,150,204}
\definecolor{DeepBlue}{RGB}{0,114,178}
\definecolor{DeepOrange}{RGB}{230,159,0}
\definecolor{TealGreen}{RGB}{0,158,115}

%% file: iclr2026/tikz_figs/fig_model_app_speedup.tex
\pgfplotsset{
  timingPlotStyle/.style={
  	tick label style={font=\normalsize},
    xmin = 2*1.5, xmax = 2^9*1.5, ymin = 0.0015, ymax = 200,
    ylabel= RNN Cell Application Timings [ms],
    log basis y={10},
    log basis x={2},
    ymajorgrids= true,
    y grid style={white},
    unbounded coords=jump,
    xlabel= Sequence length $L$,
    ymajorgrids= true,
    axis background/.style={fill=gray!10},
  }
}
\pgfplotsset{
  barPlotStyle/.style={
  	tick label style={font=\normalsize},
    xbar,
    bar width=8pt,
    xlabel={Perplexity},
    symbolic y coords={400M,1B,2.9B,7B},
    ytick=data,
    y=1.5cm, 
    enlarge y limits=0.45,
    xmin=0, xmax=15,
    axis background/.style={fill=gray!10},
    axis lines*=left,
    xmajorgrids= true,
    x grid style={white},
    nodes near coords,
    nodes near coords align={horizontal},
    nodes near coords style={
        font=\tiny,
        anchor=east,
        /pgf/number format/precision=2,
        /pgf/number format/fixed,
        text=white
    }
  }
}
\pgfplotsset{
  lineStyle/.style={
    line width=1.5pt,
    mark=*,
    mark options={solid},
    mark size=.3pt
  }
}
\tikzset{
  speedupLineStyleGru/.style={
    <->, >=stealth,
    shorten <=2pt, shorten >=2pt,
    gruColor,
    densely dotted,
    line width=.8pt
  }
}
\tikzset{
  speedupLineStyleLstm/.style={
    <->, >=stealth,
    shorten <=2pt, shorten >=2pt,
    lstmColor,
    densely dotted,
    line width=.8pt
  }
}

\begin{tikzpicture}
    \begin{loglogaxis}[name=speedup, timingPlotStyle, ytick pos=left]
        \addplot+[lineStyle, color=gruColor!40] table[x index=0, col sep=comma, restrict expr to domain={\thisrowno{0}}{1:512}, y index=1]{iclr2026/data/data_model_app_vs_seqlength.csv} coordinate[pos=1](GRUseq);\label{fig::speedup::GRU-seq}
        \addplot+[lineStyle, color=gruColor] table[x index=0, col sep=comma, restrict expr to domain={\thisrowno{0}}{1:512}, y index=4]{iclr2026/data/data_model_app_vs_seqlength.csv} coordinate[pos=1](GRUpar);\label{fig::speedup::GRU-fused}
        \addplot+[lineStyle, color=lstmColor!40] table[x index=0, col sep=comma, restrict expr to domain={\thisrowno{0}}{1:512}, y index=5]{iclr2026/data/data_model_app_vs_seqlength.csv} coordinate[pos=1](LSTMseq);\label{fig::speedup::LSTM-seq}
        \addplot+[lineStyle, color=lstmColor] table[x index=0, col sep=comma, restrict expr to domain={\thisrowno{0}}{1:512}, y index=8]{iclr2026/data/data_model_app_vs_seqlength.csv} coordinate[pos=1](LSTMpar);\label{fig::speedup::LSTM-fused}
        \draw[speedupLineStyleGru, transform canvas={xshift=-0mm}](GRUseq) -- (GRUpar) node[gruColor, midway, anchor=north east, font=\small, rotate=90, xshift=13mm, yshift=0mm] {Speedup$>665\times$};
        \draw[speedupLineStyleLstm, transform canvas={xshift=-1.5mm}](LSTMseq) -- (LSTMpar) node[lstmColor, midway, anchor=south east, font=\small, rotate=90, xshift=10.5mm, yshift=0mm] {Speedup$>447\times$};
    \end{loglogaxis}
    \matrix [ampersand replacement=\&,
            scale=0.6,
            fill=white,
            fill opacity=0.9,
            matrix of nodes,
            anchor=north west,
            inner xsep=2pt,
            shift={(10pt,-9pt)},
            nodes={font=\tiny,text width=11mm,align=center,text height=1.4mm},
           ] at (speedup.north west) {
                                        \& Sequential                    \& \method\\[-4pt]
             |[text width=11mm]|\GRUD   \& \ref*{fig::speedup::GRU-seq}  \& \ref*{fig::speedup::GRU-fused}\\[-4pt]
             |[text width=11mm]|\LSTMD  \& \ref*{fig::speedup::LSTM-seq} \& \ref*{fig::speedup::LSTM-fused}\\
        };
    \begin{axis}[name=pplplot, barPlotStyle, at={(speedup.east)}, anchor=west, xshift=2cm]
        \pgfmathsetmacro{\baseshift}{4.5}
        \addplot+[bar shift=3*\baseshift pt,fill=mambaColor,draw=mambaColor] coordinates {
            (12.257164922948800,1B)
            (10.12802277935240,2.9B)
            (8.620966296746499,7B)
        };\label{fig::ppl::mamba}
        \addplot+[bar shift=\baseshift pt,fill=lstmColor,draw=lstmColor] coordinates {
            (13.107594948023200,1B)
            (10.873804181097800,2.9B)
            (9.15825470800822,7B)
        };\label{fig::ppl::lstm}
        \addplot+[bar shift=-\baseshift pt,fill=gruColor,draw=gruColor] coordinates {
            (13.196570879517100,1B)
            (10.92374661263940,2.9B)
            (9.187267565530622,7B)
        };\label{fig::ppl::gru}
        \addplot+[bar shift=-3*\baseshift pt,fill=transformerColor,draw=transformerColor] coordinates {
            (12.898222787449000,1B)
            (10.965541661024500,2.9B)
            (9.55328208192562,7B)
        };\label{fig::ppl::trans}
    \end{axis}
    \matrix [ampersand replacement=\&,
        scale=0.6,
        fill=white,
        fill opacity=0.9,
        matrix of nodes,
        anchor=north east,
        inner xsep=2pt,
        shift={(-10pt,-9pt)},
        nodes={font=\tiny,text width=2mm,align=center,text height=1.4mm},
    ] at (pplplot.north east) {
         |[text width=12mm]|Mamba2      \& \scalebox{1}[0.5]{\ref*{fig::ppl::mamba}}\\
         |[text width=12mm]|\LSTMD      \& \scalebox{1}[0.5]{\ref*{fig::ppl::lstm}}\\
         |[text width=12mm]|\GRUD       \& \scalebox{1}[0.5]{\ref*{fig::ppl::gru}}\\
         |[text width=12mm]|Transformer \& \scalebox{1}[0.5]{\ref*{fig::ppl::trans}}\\
        };
    \node[anchor=north east,inner ysep=0, outer ysep=0, align=right] at (pplplot.north west){\thead{\normalsize Model\\\normalsize Size}};
\end{tikzpicture}

%% file: iclr2026/tikz_figs/fig_kernel_time_and_model_app_vs_seqlength.tex
\pgfplotsset{
  timingPlotStyle/.style={
  	tick label style={font=\normalsize},
    xmin = 2, xmax = 2^22*1.5, ymin = 0.005, ymax = 4.5,
    log basis y={10},
    log basis x={2},
    ymajorgrids= true,
    y grid style={white},
    unbounded coords=jump,
    xlabel= Sequence length $L$,
    ylabel= Timings [ms],
    ymajorgrids= true,
    axis background/.style={fill=gray!10},
    legend style={pos=north west,anchor=east, font=\normalsize, very thin, draw=gray},
  }
}
\pgfplotscreateplotcyclelist{timingsCycleList}{%
    {color=gruColor!33,   line width=.75pt},
    {color=lstmColor!33,  line width=.75pt},
    {color=gruColor!66,   line width=1.2pt},
    {color=lstmColor!66,  line width=1.2pt},
    {color=mambaColor!66, line width=1.2pt},
    {color=gruColor,      line width=1.75pt},
    {color=lstmColor,     line width=1.75pt}
}

\pgfplotsset{
  lineStyle/.style={
    mark=*,
    mark options={solid},
    mark size=.3pt
  }
}
\begin{tikzpicture}
    \begin{loglogaxis}[name=kerTime, timingPlotStyle, cycle list name=timingsCycleList, title=\textsc{Parallel Reduction Operations}, ytick pos=left] 
        \fill[white, opacity=0.9] (axis cs:480,0.001) rectangle (axis cs:550,100);
        \addplot+[lineStyle] table[x index=0, y index=1, col sep=comma ]{iclr2026/data/data_kernel_time_vs_seqlength.csv};\label{fig::kerTime::D-torch}
        \addplot+[lineStyle] table[x index=0, y index=3, col sep=comma ]{iclr2026/data/data_kernel_time_vs_seqlength.csv};\label{fig::kerTime::BD-torch}
        \addplot+[lineStyle] table[x index=0, y index=2, col sep=comma ]{iclr2026/data/data_kernel_time_vs_seqlength.csv};\label{fig::kerTime::D-CUDA}
        \addplot+[lineStyle] table[x index=0, y index=4, col sep=comma ]{iclr2026/data/data_kernel_time_vs_seqlength.csv};\label{fig::kerTime::BD-CUDA}
        \addplot+[lineStyle] table[x index=0, y index=5, col sep=comma ]{iclr2026/data/data_kernel_time_vs_seqlength.csv};\label{fig::kerTime::Mamba}
        \addplot+[domain=2^2:2^21, samples=50, color=black, densely dotted, line width=0.25pt, forget plot, postaction={decorate,decoration={
                markings, mark=at position 1 with {
                    \node[anchor=north, font=\tiny] {$\sim\log L$};
                }}}]{0.004*log2(x)};
    \end{loglogaxis}
    \matrix [ampersand replacement=\&,
            scale=0.6,
            fill=white,
            fill opacity=0.9,
            matrix of nodes,
            anchor=north west,
            inner xsep=2pt,
            shift={(10pt,-9pt)},
            nodes={font=\tiny,text width=7mm,align=center,text height=1.4mm},
           ] at (kerTime.north west) {
                                              \& |[text width= 9mm]|\hyperref[imp::torch]{PyTorch} \& \hyperref[imp::CUDA]{CUDA}   \\[-4pt]
             |[text width= 14mm]|Diag         \& |[text width= 9mm]|\ref*{fig::kerTime::D-torch}   \& \ref*{fig::kerTime::D-CUDA}  \\[-4pt]
             |[text width= 14mm]|BlockDiag2x2 \& |[text width= 9mm]|\ref*{fig::kerTime::BD-torch}  \& \ref*{fig::kerTime::BD-CUDA} \\
             |[text width= 14mm]|Mamba scan   \& |[text width= 9mm]|                               \& \ref*{fig::kerTime::Mamba}   \\[-4pt]
        };
    \begin{loglogaxis}[name=modApp, timingPlotStyle, cycle list name=timingsCycleList, cycle list shift=2, title=\textsc{Full RNN Cell Forward Pass}, ytick pos=right, at={(kerTime.outer east)}, anchor=outer west, xmax = 2^17*1.5] 
        \fill[white, opacity=0.9] (axis cs:480,0.001) rectangle (axis cs:550,100);
        \addplot+[lineStyle] table[x index=0, y index=3, col sep=comma ]{iclr2026/data/data_model_app_vs_seqlength.csv};\label{fig::modApp::GRU-CUDA}
        \addplot+[lineStyle] table[x index=0, y index=7, col sep=comma ]{iclr2026/data/data_model_app_vs_seqlength.csv};\label{fig::modApp::LSTM-CUDA}
        \addplot+[lineStyle] table[x index=0, y index=9, col sep=comma ]{iclr2026/data/data_model_app_vs_seqlength.csv};\label{fig::modApp::Mamba}
        \addplot+[lineStyle] table[x index=0, y index=4, col sep=comma ]{iclr2026/data/data_model_app_vs_seqlength.csv};\label{fig::modApp::GRU-fused}
        \addplot+[lineStyle] table[x index=0, y index=8, col sep=comma ]{iclr2026/data/data_model_app_vs_seqlength.csv};\label{fig::modApp::LSTM-fused}
    \end{loglogaxis}
    \matrix [ampersand replacement=\&,
            scale=0.6,
            fill=white,
            fill opacity=0.9,
            matrix of nodes,
            anchor=north west,
            inner xsep=2pt,
            shift={(10pt,-9pt)},
            nodes={font=\tiny,text width=7mm,align=center,text height=1.4mm},
           ] at (modApp.north west) {
                                       \& \hyperref[imp::CUDA]{CUDA}    \& \hyperref[imp::CUDA-fused]{Fused}\\[-4pt]
             |[text width=11mm]|\GRUD  \& \ref*{fig::modApp::GRU-CUDA}  \& \ref*{fig::modApp::GRU-fused}  \\[-4pt]
             |[text width=11mm]|\LSTMD \& \ref*{fig::modApp::LSTM-CUDA} \& \ref*{fig::modApp::LSTM-fused} \\
             Mamba  \& \ref*{fig::modApp::Mamba}     \\[-4pt]
        };
\end{tikzpicture}

%% file: iclr2026/tikz_figs/fig_model_throughput.tex
\pgfplotsset{
  throughputPlotStyle/.style={
  	tick label style={font=\normalsize},
    xmin = 0.6, xmax = 2^11*1.5, ymin = 1, ymax = 100,
    log basis y={10},
    log basis x={2},
    ymajorgrids= true,
    y grid style={white},
    unbounded coords=jump,
    xlabel= Number of new tokens generated $L$,
    ylabel= Throughput [Tokens/s],
    ymajorgrids= true,
    axis background/.style={fill=gray!10},
    cycle list = {{gruColor},{lstmColor},{mambaColor},{transformerColor}},
    width=240pt,
    height=170pt
  }
}
\pgfplotsset{
  lineStyle/.style={
    line width=0.4mm,
    mark=*,
    mark options={solid},
    mark size=.3pt
  }
}
\begin{tikzpicture}
    \begin{loglogaxis}[name=modelThroughput, throughputPlotStyle, title=\textsc{Model Throughput}, ytick pos=left] 
        \addplot+[lineStyle] table[x index=0, y index=5,  col sep=comma ]{iclr2026/data/data_throughput_timings.csv};\label{fig::throughput::GRU}
        \addplot+[lineStyle] table[x index=0, y index=7,  col sep=comma ]{iclr2026/data/data_throughput_timings.csv};\label{fig::throughput::LSTM}
        \addplot+[lineStyle] table[x index=0, y index=9,  col sep=comma ]{iclr2026/data/data_throughput_timings.csv};\label{fig::throughput::Mamba}
        \addplot+[lineStyle] table[x index=0, y index=11, col sep=comma ]{iclr2026/data/data_throughput_timings.csv};\label{fig::throughput::Trans}
        \addplot+[domain=2^8:2^11, samples=50, color=black, line width=0.25pt, densely dotted, forget plot, postaction={decorate,decoration={
                markings, mark=at position 0.7 with {
                    \node[anchor=south west, font=\tiny] {$\sim \frac{1}{L}$};
                }}}]{4300/x};
    \end{loglogaxis}
    \matrix [ampersand replacement=\&,
            scale=0.6,
            fill=white,
            fill opacity=0.9,
            matrix of nodes,
            anchor=south west,
            inner xsep=2pt,
            shift={(10pt,9pt)},
            nodes={font=\tiny,text width=13mm,align=center,text height=1.4mm},
           ] at (modelThroughput.south west) {
             \GRUD       \& |[text width= 9mm]|\ref*{fig::throughput::GRU}   \&$\sim37\text{ tkn/s}$\\[-4pt]
             \LSTMD      \& |[text width= 9mm]|\ref*{fig::throughput::LSTM}  \&$\sim35\text{ tkn/s}$\\[-4pt]
             Mamba2      \& |[text width= 9mm]|\ref*{fig::throughput::Mamba} \&$\sim27\text{ tkn/s}$\\[-4pt]
             Transformer \& |[text width= 9mm]|\ref*{fig::throughput::Trans} \&\\[-4pt]
        };
\end{tikzpicture}

%% file: apple_template/appendix.tex
\section{Newton convergence for \GRUD and \LSTMD}
\label{app::newton}

\begin{summarybox}
In this section, we empirically verify that \method's Newton algorithm is stable and efficient when applied to \GRUD and \LSTMD, consistently converging in 3 iterations, regardless of RNN cell used and stage of training considered.
\end{summarybox}

In \cref{sec::limitations} we hinted at the necessity of the Newton's method \cref{alg::newton} to converge quickly for \method to be appealing as a parallelization method for RNN cells applications. In practice, solving \cref{eqn::state} in parallel comes at a computational complexity of $\mc{O}(N_{\text{its}}\log_2L)$, if $N_{\text{its}}$ is the number of Newton iterations required for convergence: keeping $N_{\text{its}}\sim\mc{O}(1)$ is then paramount. Ultimately, the convergence behavior of Newton's method will depend on the actual definition of the cell action $\bb{f}$ in \cref{eqn::rnn_seq}, and providing a comprehensive theoretical study of the properties of $\bb{f}$ necessary for fast convergence is beyond the scope of this project and a topic for future research---although we refer to \cite{gonzalez2025predictabilityenablesparallelizationnonlinear} for the relevance of Lyapunov stability in this sense. Here, we limit ourselves to demonstrating empirically that, when applied to the \GRUD and \LSTMD models considered in this work, Newton's method consistently converges around the third iteration.

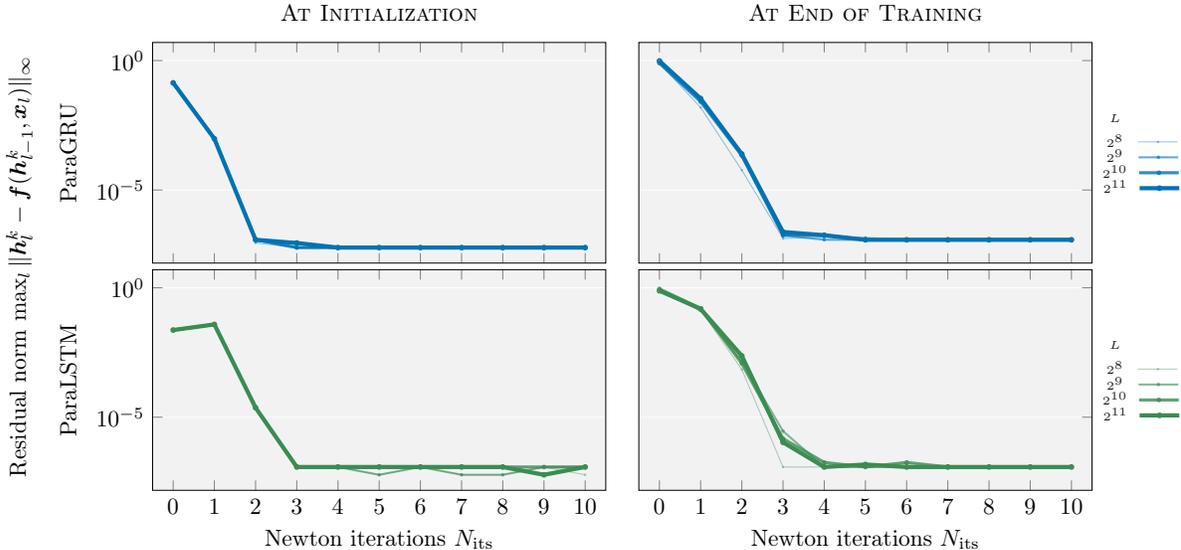
\begin{figure}[b!]
    \centering
    \vskip 0.2in
    \resizebox{0.96\textwidth}{!}{\input{iclr2026/tikz_figs/fig_newton_conv}}
    \caption{Newton's method convergence behavior for \GRUD (top) and \LSTMD (bottom) cells with different input sequence lengths $L$, for freshly initialized RNN cells (left), and for the RNN cells in the last layer of our trained 400M models (right). Inputs are randomly selected from the evaluation datasets, and the residuals reported are the $\max$ on batches of size 8.}
    \label{fig::newton_convergence}
    \vskip -0.2in
\end{figure}

To this end, in \cref{fig::newton_convergence} we examine the convergence behavior of Newton's method across different stages of model training and for varying sequence lengths. By default, in all forward applications of any RNN cell (including during the training of the models described in \cref{sec::performance}), we consider as an initial guess $\bb{h}^0$ for the Newton iterations the quantity
\begin{equation}
    \bb{h}_{l}^0 = \bb{f}(\bb{0},\bb{x}_l),\qquad\qquad\forall l=1,\dots,L.
\end{equation}
That is, for each input $\bb{x}_l$, we apply the cell action considering an all-zero previous hidden state $\bb{0}$. Notice this is computed in a perfectly parallel fashion along the sequence length $L$.

\Cref{fig::newton_convergence} considers two complementary regimes, to show that Newton iterations are stable throughout the training procedure. On the left we show residual norm convergence for freshly-initialized RNN cells (see \cref{app::llm} for the initialization scheme considered), while on the right we consider RNN cells extracted from the last layer of the fully-trained 400M-parameter models described in \cref{app::llm}. As evident from the graphs, the method exhibits rapid convergence: the residual drops to machine precision within 3-4 iterations, regardless of RNN cell considered or training stage, and convergence rate is only very marginally affected by an increase in the sequence length. For freshly initialized \LSTMD, the residual at the very first iteration increases, indicating potential marginal for improvement in defining an initial guess for Newton's method. For \GRUD, just 2 iterations suffice at initialization, growing to 3 at the end of training. These results guided our choice of $N_\text{its}=3$ for the experiments in \cref{sec::performance}.

This consistent behavior across both initialization and trained states suggests that the nonlinearity structure of the \GRUD and \LSTMD cells lends itself well to a solution via Newton's method, but---we remark---this is not in general guaranteed for every choice of RNN cell.

\clearpage
\section{Details on profiling results}
\label{app::profiling}



\begin{summarybox}
In this section we expand the results in \cref{sec::efficiency} reporting timings for all application modalities of the RNN cells, including sequential and pure PyTorch implementations. The fully-fused forward pass applications of \method are always faster or comparable to Mamba for all the sequence lengths tested, and beat Mamba consistently when the backward pass is also taken into account.
\\
\\
Additionally we present an ablation over the \texttt{chunk\_size} hyperparameter, controlling thread-level workload in our CUDA implementations. The resulting optimal values are model- and application mode-specific, with values $1\sim4$ generally yielding best runtimes.
\end{summarybox}

\paragraph{Experiments Setup}
All timings in \cref{sec::efficiency} are collected on an NVIDIA A100 GPU. For \cref{fig::kernel_and_model_own_vs_mamba}, profiling is performed by tracking CUDA \texttt{Event}s \cite[Sec.~6.2.8.8.2.]{cuda2025}. The profiler first warms-up the GPU with 20 dry runs of the target kernel to measure, before proceeding to collecting measurements from 100 runs processing random input sequences of a given length $L$. In the figures we report the minimum\footnote{We report the \emph{minimum} following established benchmarking practices in high-performance computing: it represents a cleaner metric, and a more faithful indication of the actual performance of a kernel operating under ideal conditions. The \emph{average}, on the other hand, might be polluted by transient system interference (e.g., synchronization artifacts, GPU thermal throttling events, OS-induced delays, ...) unrelated to algorithmic efficiency.} timings for each sequence length considered. To provide a fairer comparison against production-ready code, all PyTorch baseline measurements (including Mamba) are performed using \texttt{torch.compile()}-optimized code, rather than eager execution mode, as this represents the standard deployment configuration for real-world applications. Also for the sake of fairness, for the Mamba baselines in \cref{fig::kernel_and_model_own_vs_mamba} we consider: on the left, the application of the sole \texttt{parallel\_scan} routine at the core of Mamba's SSM application; on the right, the application of a \emph{simplified} Mamba module, stripped of its gate and convolution components \citep{mamba}---in other words, we track only those operations necessary for the assembly of the SSM parameters and the application of the SSM cell. Finally, full RNN cell application timings consider the default setup with $N_{\text{its}}=3$ Newton iterations.

For the throughput results in \cref{fig::throughput_own_vs_mamba}, we consider the 1B models discussed in \cref{sec::performance}. To profile, we simply use Python's \texttt{timeit} module, and report the minimum timings over 5 repetitions. The reduced number of repetitions in this case is mainly due to the lengthy generation times required by the Transformer; nonetheless, the reported curves appear smooth, so we consider them sufficient to provide an accurate measure.

\subsection{Additional timing results}
\begin{figure}[t!]
    \centering
    \vskip 0.2in
    \resizebox{0.96\textwidth}{!}{\input{iclr2026/tikz_figs/fig_model_own_vs_mamba_zoom}}
    \caption{Timings of different application modalities for the \GRUD (left) and \LSTMD (right) RNN cells, applied to sequences of varying length. Heavier lines refer to progressively more efficient applications: sequential, using parallel reduction in PyTorch (\cref{alg::newton}), using parallel reduction in CUDA and system assembly in PyTorch, and using a fully-fused CUDA kernel for the whole Newton routine. Equivalent timing results for Mamba are also included for reference (yellow). Same setup as for \cref{fig::kernel_and_model_own_vs_mamba} (right).}
    \label{fig::model_own_vs_mamba_zoom}
    \vskip -0.2in
\end{figure}
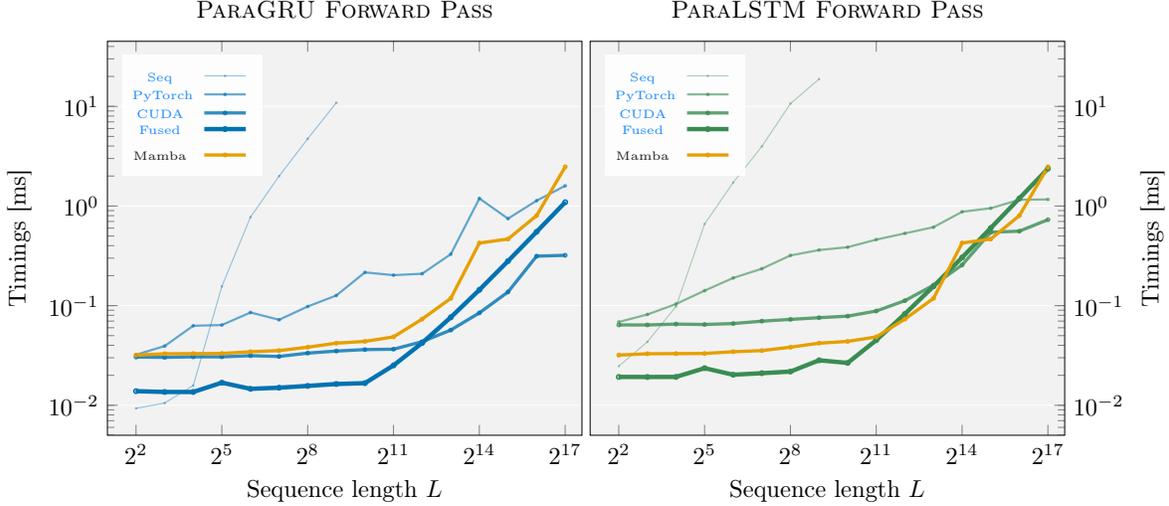
\paragraph{Forward / Backward Pass Varying Application Modality} In \cref{fig::model_own_vs_mamba_zoom} we expand on the results reported in \cref{fig::kernel_and_model_own_vs_mamba}, including timings for the sequential application of the RNN cells, as well as from the pure PyTorch implementation. The former results were also used to compute the speedup measurements reported in \cref{fig::speedup}. Overall, the curves behave as expected: applying the RNN cells sequentially becomes quickly computationally unfeasible due to its linear-growth complexity, and moving to progressively more efficient implementations (from sequential, to parallel in PyTorch, to computing the parallel reduction operation in CUDA, to a fully-fused CUDA kernel for the whole Newton routine) lowers the overall runtime for the RNN application. Perhaps surprisingly, the curve for the PyTorch implementation of \GRUD in \cref{fig::model_own_vs_mamba_zoom} (left) is ragged. We speculate that this is due to the effect of \texttt{torch.compile()}, which for this cell is able to unlock certain optimizations for specific sequence lengths but not for others. Notice also that the CUDA implementation of \GRUD surpasses the fully-fused one for $L>2^{12}$. To explain this unexpected result we need to refer to the specifics of the CUDA kernel implementation outlined in \cref{app::code::solver}. Fully-fused kernels exert a larger pressure on the GPU registers, in light of the larger number of computations they must perform (Jacobian assembly and Newton iterations, on top of parallel reduction). To ensure that a GPU block has enough resources to carry these computations, then, we are forced to reduce the \texttt{chunk\_size} hyperparameter which controls the number of equations a single thread processes. This in turn triggers the block-sequential solution (hence the collapse to linear regime) for shorter sequence lengths than for the CUDA implementation ($L>2^{10}$ instead of $L>2^{11}$), thus explaining the behavior of the curves in \cref{fig::model_own_vs_mamba_zoom} (left).

\begin{figure}[b!]
    \centering
    \vskip 0.2in
    \resizebox{0.96\textwidth}{!}{\input{iclr2026/tikz_figs/fig_model_own_vs_mamba_zoom_fwdbwd}}
    \caption{Timings of different application modalities for a full forward and backward pass of the \GRUD (left) and \LSTMD (right) RNN cells, applied to sequences of varying length. Same setup as for \cref{fig::model_own_vs_mamba_zoom}.}
    \label{fig::model_own_vs_mamba_zoom_fwdbwd}
    \vskip -0.2in
\end{figure}
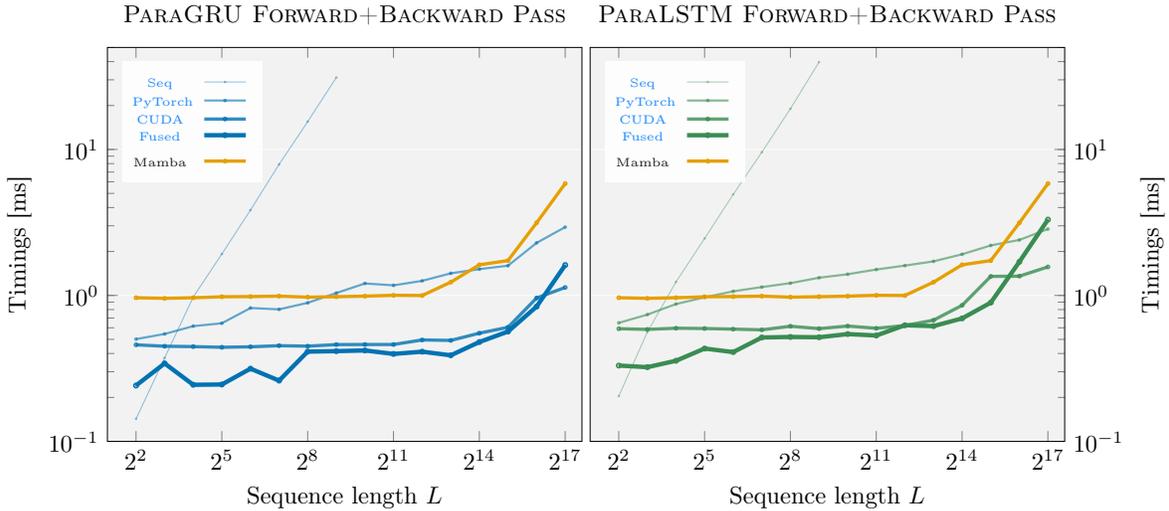    

In \cref{fig::model_own_vs_mamba_zoom_fwdbwd} we further provide timings for a full parallel forward and backward application of our RNN cells, using a dummy loss function $\mc{L}(\bb{h}) = \sum_l\bb{h}_l^2$. The setup is analogous to that for \cref{fig::model_own_vs_mamba_zoom}, and the observations regarding relative performance of the various implementations remain substantially unchanged. Comparing with \cref{fig::model_own_vs_mamba_zoom}, however, we can notice that including the backward pass affects the timings for Mamba much more than that for \GRUD or \LSTMD. This is in line with expectations: in Mamba, both forward and backward passes involve the application of a \emph{single} parallel reduction operation \cref{alg::pcr}. On the other hand, for nonlinear RNNs like \GRUD or \LSTMD, the forward pass requires \emph{multiple} parallel reductions---one per each Newton iteration, that is $N_\text{its}=3$ in our case---while the backward pass only requires a \emph{single} parallel reduction, as outlined in \cref{sec::method}.

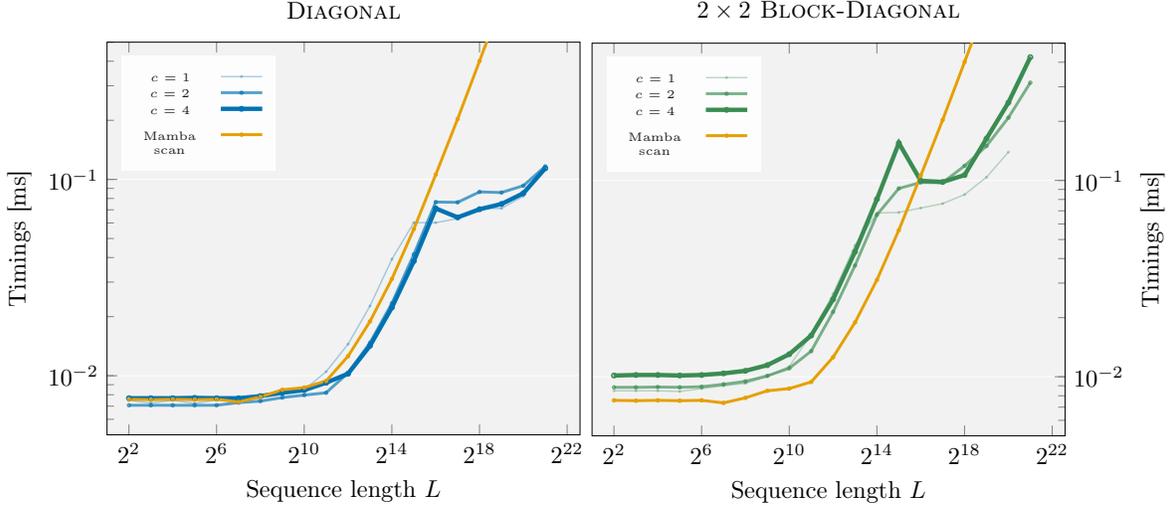
\begin{figure}[t!]
    \centering
    \vskip 0.2in
    \resizebox{0.96\textwidth}{!}{\input{iclr2026/tikz_figs/fig_kernel_time_vs_chunksize}}
    \caption{Timings of the CUDA implementation of parallel reduction, for diagonal (left) and $2\times2$ block-diagonal (right) Jacobians, varying its \texttt{chunk\_size} hyperparameter $c$. Timings for Mamba implementation of \texttt{parallel\_scan} are also reported as reference.}
    \label{fig::kernel_time_vs_chunksize}
    \vskip -0.2in
\end{figure}    
\begin{figure}[b!]
    \centering
    \vskip 0.2in
    \resizebox{0.96\textwidth}{!}{\input{iclr2026/tikz_figs/fig_model_app_vs_chunksize}}
    \caption{Timings of the implementation of the fully-fused kernel for parallel RNN application, for \GRUD (left) and \LSTMD (right), varying its \texttt{chunk\_size} hyperparameter $c$. Timings for Mamba SSM application are also reported as reference.}
    \label{fig::model_app_vs_chunksize}
    \vskip -0.2in
\end{figure}
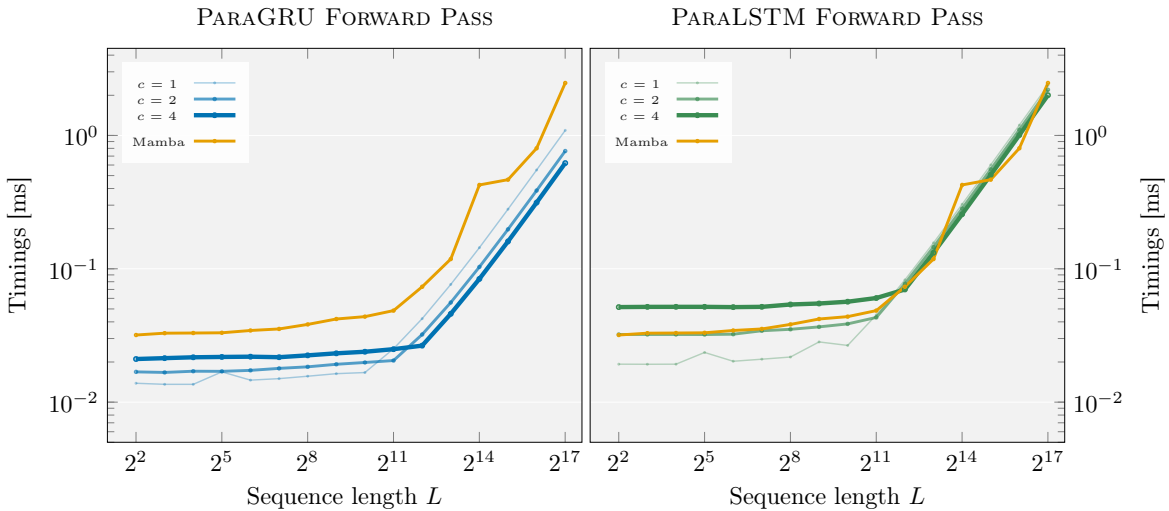    

\paragraph{Ablation Over Kernel \texttt{chunk\_size} Hyperparameter} In \cref{app::code::solver::cuda} we comment on the various compile-time hyperparameters one can tune to achieve the highest performance out of \method for their target setup. Likely the most relevant one is \texttt{chunk\_size}, which defines the amount of work per thread---that is, the number of equations each thread must solve individually. Here we report the results of the sweep we conducted with the goal of selecting its optimal value. In particular, in \cref{fig::kernel_time_vs_chunksize} we show timings for the sole parallel reduction operation applied to diagonal (left, used by \GRUD) and $2\times2$ block-diagonal Jacobians (right, used by \LSTMD), when varying \texttt{chunk\_size}. For the diagonal case, we pick a value of $c=2$ as the optimal, as it achieves the best performance for $L\lesssim2^{12}$, but $c=4$ would perform marginally better for $L\gtrsim2^{12}$. Also for the block-diagonal counterpart, we consider $c=2$, as it performs the best in the regime relevant to our training $2^{10}\lesssim L\lesssim2^{14}$, although for shorter sentences $c=1$ might be preferable.

We conduct a similar sweep also for the fully-fused parallel application of \GRUD and \LSTMD, and report its results in \cref{fig::model_app_vs_chunksize}. For \GRUD, the main advantage in increasing $c>1$ consists in delaying the triggering of the linear regime, and indeed the choice of $c=2$ is optimal for the $L=2^{11}$ used in our training; for shorter sequences, however, sticking to $c=1$ gives better runtimes. For \LSTMD, this advantage is instead negligible, and using $c=1$ throughout gives the best runtimes overall.

\newpage
\section{Details on language modeling results}

\begin{summarybox}
In this section we provide more information about the architectures we used, the training parameters, timing protocol, and results regarding the language modeling experiments. Among those results:
\begin{itemize}[leftmargin=.5cm]
\item  In \cref{tab::lm-eval-harness_complete} we expand upon the results illustrated in \cref{tab::lm-eval-harness}, providing the downstream tasks evaluation scores across all model scales considered (400M, 1B, 2.9B, and 7B). Performance remains consistent across different parameter counts: with Mamba2, \GRUD and \LSTMD often outperforming the Transformer. Importantly, our RNN cells show consistent improvement with scale, and the relative performance gaps between architectures remain stable.
\item In \cref{tab::full_synthetic} we expand the results shown in \cref{tab::synthetic} by adding more tasks: these further confirm the role of nonlinearities in boosting expressivity.
\end{itemize}
\end{summarybox}

\subsection{Language Modeling}
\label{app::llm}

\paragraph{Models Description}
\begin{wrapfigure}[23]{r}{0.55\textwidth}
    \centering
    \includegraphics[trim={375mm 25mm 375mm 25mm},clip,width=0.4\textwidth]{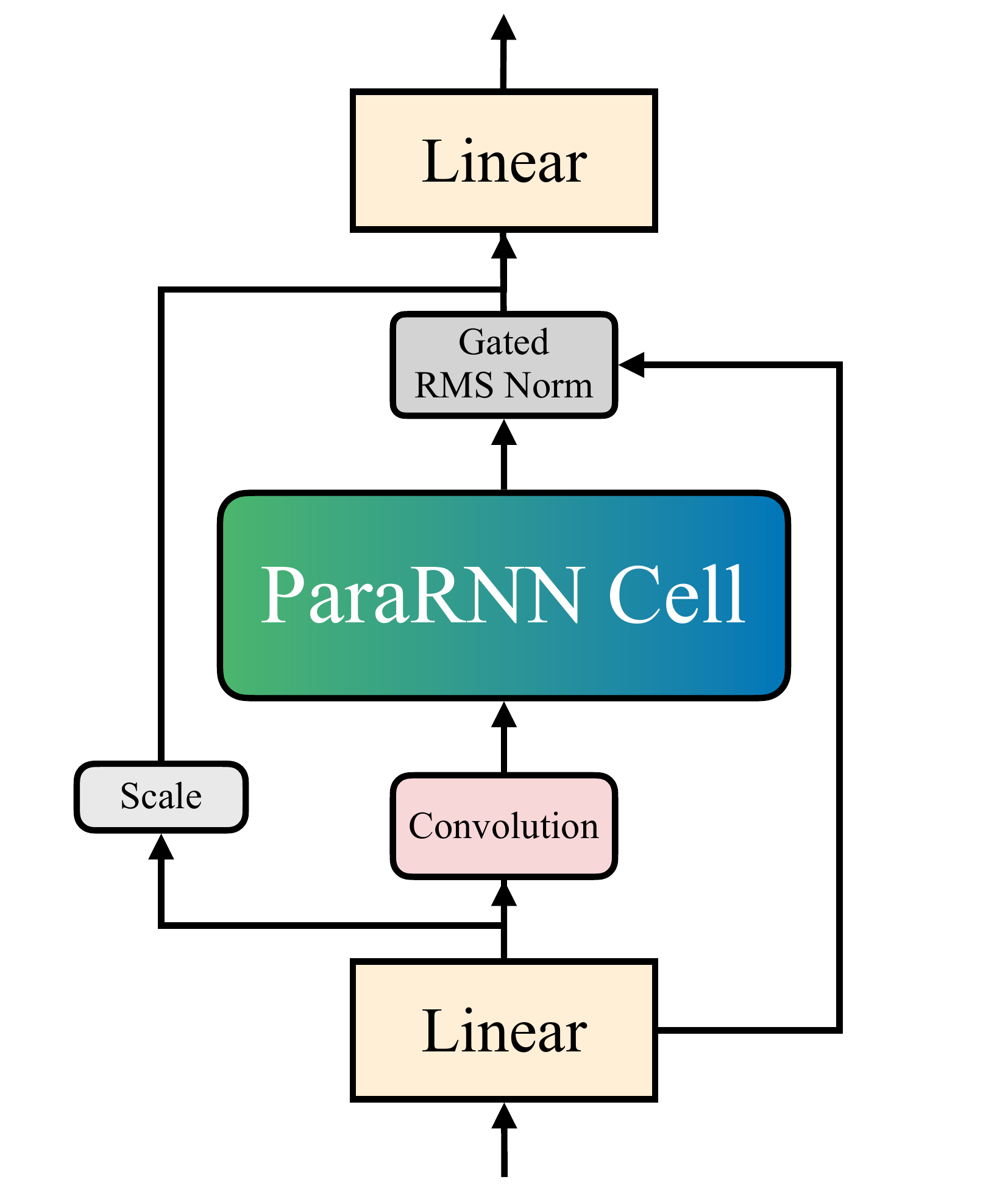}
    \caption{Schematics of RNN block.}
    \label{fig::RNN_model_schematics}
\end{wrapfigure}
Schematics of the complete RNN block used for the language modeling experiments in \cref{sec::results} are shown in \cref{fig::RNN_model_schematics}. The module inherits some components from Mamba (a pre-pending short sequence-wise causal convolution, as well as a gated RMS normalization layer following the actual RNN cell application), and includes a residual connection with a learnable scaling of the input. Still, our block follows closely the Transformer structure: particularly, MLP layers with residual connections are interwoven with the sequence mixer, which in our case is an RNN cell, rather than Attention. The core RNN cell is either \GRUD or \LSTMD from \cref{sec::LSTMD}, but all other architectural components remain identical between the two variants. To aid with scaling, we use a multi-head implementation of our RNN cells, where the input $\bb{x}_l\in\mathbb{R}^{d_{\text{in}}}$ is separated into $n_{\text{heads}}$ independent heads, each of size $d_h = d_{\text{in}}/n_{\text{heads}}$. The whole block is then repeated $n_{\text{layers}}$ times to generate the full model.

The baselines considered in our experiments are the Transformer architecture in DCLM~\citep{li2024datacomplm}, and the Mamba2 model from HuggingFace. See \cref{tab::hparams} for a summary of hyperparameters further detailing the architectures used.

\begin{table}[t!]
  \small
  \centering
  \caption{Architecture specifications and optimal hyperparameters for our experiments. We report structural parameters (number of layers, number of heads and model width) and training hyperparameters (learning rate and weight decay) that achieved lowest perplexity for each model type and scale. Time measurements show average single-step training cost on NVIDIA H100 GPUs.}
  \label{tab::hparams}
  \begin{tabular}{clccccccc}
    \toprule
    & Model type & \# params & $n_{\text{layers}}$ & $n_{\text{heads}}$ & $d_{\text{model}}$ & \thead{Learning\\rate} & \thead{Weight\\decay} & \thead{$\downarrow$Time\\{[ms/step]}} \\
    \midrule
    \multirow{4}{*}{\rotatebox{90}{400M}}
    & \GRUD & 406M & 24 & 4 & 1024 & 0.005 & 1e-3 & 95.7 \\
    & \LSTMD & 406M & 24 & 4 & 1024 & 0.003 & 1e-3 & 116 \\
    & Mamba2 & 409M & 48 & 32 & 1024 & 0.005 & 1e-3 & 115 \\
    & Transformer & 412M & 24 & 8 & 1024 & 0.003 & 1e-3 & 81.0 \\
    \midrule
    \multirow{4}{*}{\rotatebox{90}{1B}}
    & \GRUD & 1.42B & 24 & 4 & 2048 & 0.003 & 1e-4 & 192 \\
    & \LSTMD & 1.42B & 24 & 4 & 2048 & 0.003 & 1e-4 & 235 \\
    & Mamba2 & 1.43B & 48 & 64 & 2048 & 0.005 & 1e-4 & 221 \\
    & Transformer & 1.44B & 24 & 16 & 2048 & 0.002 & 1e-4 & 157 \\
    \midrule
    \multirow{4}{*}{\rotatebox{90}{2.9B}}
    & \GRUD & 2.90B & 32 & 2 & 2560 & 0.002 & 1e-4 & 326 \\
    & \LSTMD & 2.90B & 32 & 2 & 2560 & 0.002 & 1e-4 & 400 \\
    & Mamba2 & 2.83B & 64 & 80 & 2560 & 0.003 & 1e-4 & 393 \\
    & Transformer & 2.80B & 32 & 32 & 2560 & 0.003 & 1e-4 & 267 \\
    \midrule
    \multirow{4}{*}{\rotatebox{90}{7B}}
    & \GRUD & 6.76B & 32 & 4 & 4096 & 0.003 & 1e-4 & 613 \\
    & \LSTMD & 6.76B & 32 & 4 & 4096 & 0.003 & 1e-4 & 715 \\
    & Mamba2 & 6.96B & 64 & 128 & 4096 & 0.003 & 1e-4 & 9475 \\
    & Transformer & 6.89B & 32 & 32 & 4096 & 0.003 & 1e-4 & 520 \\
    \bottomrule
  \end{tabular}
\end{table}

\begin{table}[b!]
\small
\centering
\caption{Scale-specific training configurations on nodes with 8 NVIDIA H100 GPUs each All models use 2048-token sequences and follow ($\sim 1\times$) Chinchilla-optimal token budgets~\citep{hoffmann2022trainingcomputeoptimallargelanguage} for batch-size selection. The z-loss regularization coefficient is taken from DCLM~\citep{li2024datacomplm}.}
\label{tab::training_config}
\begin{tabular}{lccccccc}
\toprule
    \thead{Model\\Scale} & Nodes & \thead{Batch Size\\per GPU} & \thead{Effective\\Batch Size} & Z-Loss & \thead{Total\\Iterations} & \thead{Warm-up\\Iterations} & \thead{Training\\Tokens [B]} \\
\midrule
400M & 1 & 32 & 256 & 1e-04 & 20,000 & 2,000 & 10.5 \\
1B & 1 & 32 & 256 & 1e-04 & 55,000 & 5,500 & 28.8 \\
2.9B & 4 & 16 & 512 & 1e-04 & 55,000 & 5,500 & 57.7 \\
7B & 32 & 8 & 2,048 & 5e-06 & 33,000 & 3,300 & 138.4 \\
\bottomrule
\end{tabular}
\end{table}

\paragraph{Training Setup}
For our experiments we consider four architecture types---the \GRUD and \LSTMD introduced in this work, plus Mamba2 and Transformer as baselines: see \cref{tab::hparams} for architecture specifics---each at four different scales, of 400M, 1B, 2.9B and 7B parameters. As training dataset, we use SlimPajama (SPJ) \citep{cerebras2023slimpajama}, from which we removed the Books3 split (as it is known to contain copyrighted material). All models are trained using AdamW with detached weight decay and a cosine learning rate schedule, incorporating a warm-up for 10\% of the total training iterations and decaying to a final learning rate of 0~\citep{bergsmastraight}. Training is conducted on nodes with 8 NVIDIA H100 GPUs each using PyTorch 2.6 with FSDP and native automatic mixed precision, storing weights in \texttt{bfloat16} while maintaining gradients and reduction operations in \texttt{float32}. Batch sizes and total number of training tokens follow Chinchilla scaling laws~\citep{hoffmann2022trainingcomputeoptimallargelanguage}, and we refer to \cref{tab::training_config} for the complete training configuration. Hyperparameter selection is conducted by sweeping over three learning rate and weight decay values for each model and each size, selecting the configuration resulting in lowest perplexity on the SPJ test set (see \cref{tab::hparams} for the optimal values selected in our runs). Model initialization follows architecture-specific strategies: for Mamba2, we use the default initialization available in HuggingFace. For Transformers, we use DCLM initialization~\citep{li2024datacomplm}. Since our \GRUD and \LSTMD blocks in \cref{fig::RNN_model_schematics} mimic the Transformer architecture, we opt for DCLM initialization for them as well, except for their core RNN cells, whose initialization depends on the parameter considered: for the cell input matrices $B_\star$ in \cref{eqn::GRU_LSTM} we use \textit{Kaiming Uniform}~\citep{kaiminguniform} initialization; for the (diagonal) state and peephole matrices $\bb{a}_\star$ and $\bb{c}_\star$ in \cref{eqn::GRU_LSTM_simp} we choose \textit{Xavier Gaussian}, since we found small values improved stability; the bias terms $\bb{b}_\star$ are instead initialized to $\bb{0}$. We also clip the norm of $\bb{a}_\star$ and $\bb{c}_\star$ to a maximum of $0.5$ to prevent overflows in the computation of the hidden state for long sequences: this effectively bounds the Jacobians in \cref{eqn::GRU_LSTM_jac}, similarly to what is done in Mamba, where the state matrix eigenvalues are constrained in $(0,1)$.

\begin{figure}[b!]
    \centering
    \vskip 0.2in
    \includegraphics[width=0.96\textwidth]{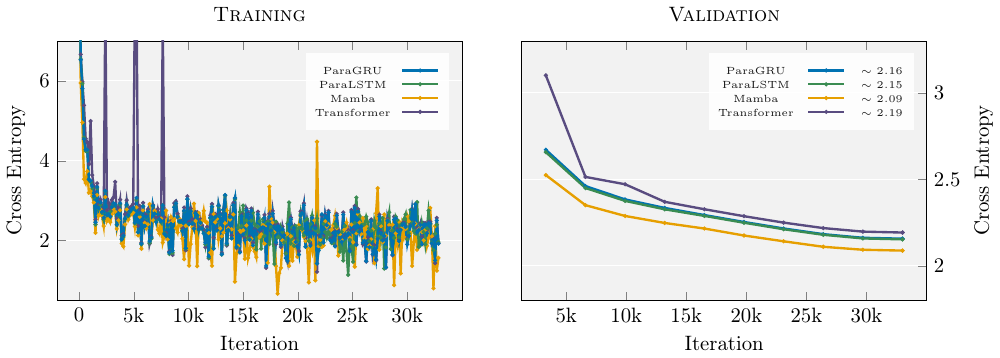}
    \caption{Cross-entropy loss evolution for the 7B models considered, evaluated on the training (left, collected every 150 iterations), and validation (right, collected every 3.3k iterations) datasets.}
    \label{fig::loss_val_train}
\end{figure}    

To illustrate the stability of our training procedure, \cref{fig::loss_val_train} shows training and validation loss curves for our 7B-parameter models. Notably, both \GRUD and \LSTMD exhibit smooth convergence throughout training---especially when compared to the baselines---suggesting that the gradient flow in our recurrent models is well-behaved at scale.

\begin{table*}[t!]
\scriptsize
\centering
\caption{Final perplexities, and evaluation scores on reference downstream tasks from \texttt{lm-eval-harness} \citep{eval-harness}, for all model types and scales considered.}
\label{tab::lm-eval-harness_complete}
\resizebox{\textwidth}{!}{
\begin{tabular}{ll cc c@{\hspace{4pt}}c c@{\hspace{4pt}}c c@{\hspace{4pt}}c c@{\hspace{4pt}}c cc }
    \toprule
    & & & & \multicolumn{2}{c}{$\uparrow$~Arc-C} & \multicolumn{2}{c}{$\uparrow$~HSwag} & \multicolumn{2}{c}{$\uparrow$~OBQA} & \multicolumn{2}{c}{$\uparrow$~WinoG} & $\uparrow$~PiQA & $\uparrow$~MMLU \\
    \multicolumn{2}{c}{Model} & \#params & $\downarrow$~PPL & (25) & (3) & (10) & (0) & (10) & (0) & (5) & (0) &  (0) &  (0) \\
    \midrule
    \multirow{4}{*}{\rotatebox{90}{400M}}
    &Transformer & 412M & 17.00 & 24.23 & 23.81 & 36.11 & 35.91 & 26.60 & 29.40 & 51.77 & 51.69 & 64.64 & 25.37 \\
    &\GRUD       & 406M & 18.56 & 22.27 & 21.93 & 38.69 & 39.21 & 25.00 & 25.80 & 53.04 & 52.41 & 67.30 & 23.68 \\
    &\LSTMD      & 406M & 18.40 & 23.12 & 24.23 & 37.43 & 37.52 & 26.20 & 26.20 & 51.22 & 49.49 & 65.94 & 23.37 \\
    &Mamba2      & 409M & 16.49 & 24.32 & 25.09 & 40.53 & 41.00 & 30.00 & 34.00 & 50.99 & 51.14 & 66.97 & 23.41 \\
    \midrule
    \multirow{4}{*}{\rotatebox{90}{1B}}
    &Transformer & 1.44B & 12.90 & 26.87 & 25.51 & 47.25 & 47.14 & 34.00 & 33.40 & 54.06 & 54.22 & 68.49 & 23.12 \\
    &\GRUD       & 1.42B & 13.20 & 24.15 & 23.81 & 48.71 & 48.85 & 24.80 & 25.80 & 53.43 & 53.28 & 71.27 & 24.55 \\
    &\LSTMD      & 1.42B & 13.11 & 24.15 & 24.32 & 45.82 & 46.53 & 24.40 & 27.00 & 51.14 & 51.07 & 70.46 & 24.75 \\
    &Mamba2      & 1.43B & 12.26 & 28.41 & 27.47 & 50.90 & 51.21 & 36.20 & 36.00 & 55.25 & 55.17 & 71.71 & 24.39 \\
    \midrule
    \multirow{4}{*}{\rotatebox{90}{2.9B}}
    &Transformer & 2.80B & 10.97 & 30.72 & 29.52 & 55.59 & 55.20 & 36.80 & 35.40 & 57.06 & 58.41 & 73.12  & 23.12  \\
    &\GRUD       & 2.90B & 10.92 & 23.98 & 23.12 & 57.58 & 57.94 & 26.00 & 26.60 & 53.83 & 56.51 & 74.48  & 23.75 \\
    &\LSTMD      & 2.90B & 10.87 & 24.15 & 24.06 & 54.22 & 55.14 & 26.80 & 27.40 & 53.83 & 53.75 & 71.71  & 23.99  \\
    &Mamba2      & 2.83B & 10.13 & 34.47 & 33.70 & 61.13 & 61.12 & 42.20 & 37.80 & 59.43 & 59.51 & 73.78  & 24.89  \\
    \midrule
    \multirow{4}{*}{\rotatebox{90}{7B}}
    &Transformer & 6.89B & 9.55 & 34.30 & 33.36 & 62.98 & 62.20 & 40.00 & 37.20 & 61.48 & 60.85 & 74.97  & 23.12  \\
    &\GRUD       & 6.76B & 9.19 & 39.68 & 36.77 & 65.85 & 65.75 & 42.20 & 40.40 & 61.40 & 59.83 & 76.66  & 25.29 \\
    &\LSTMD      & 6.76B & 9.16 & 37.46 & 36.52 & 62.47 & 62.85 & 42.20 & 38.80 & 57.70 & 59.12 & 75.19  & 25.31  \\
    &Mamba2      & 6.96B & 8.62 & 40.02 & 39.59 & 69.78 & 69.68 & 42.20 & 42.20 & 65.19 & 63.77 & 76.66  & 26.61  \\ 
    \bottomrule
\end{tabular}}
\end{table*}

\paragraph{Additional Results} In \cref{tab::lm-eval-harness_complete} we expand upon the results illustrated in \cref{tab::lm-eval-harness}, providing the downstream tasks evaluation scores across all model scales considered (400M, 1B, 2.9B, and 7B). Overall, the performance observed is consistent across different parameter counts: while Mamba2 still achieves the lowest perplexity, \GRUD and \LSTMD remain competitive, with all three recurrent architectures often outperforming the Transformer. Importantly, our RNN cells show consistent improvement with scale, and the relative performance gaps between architectures remain stable.

\subsection{Synthetic tasks}
\label{app::synthetic}
For the synthetic tasks in \cref{sec::performance}, we trained a \textit{single layer} of \GRUD, \LSTMD, Mamba2, and Transformer across a variety of synthetic tasks that have been used in the literature to evaluate the expressivity of sequential models. A summary of key results is shown in \cref{tab::synthetic}, and full results are shown in \cref{tab::full_synthetic}.
The evaluation tasks considered are:
\begin{description}
    \item[Keep-n\textsuperscript{th}]  Given an input sequence $[x_l]_{l=1}^{L}$ of values from a dictionary $x_l\in V$, output its $n$-th element, $x_n$.
    \item[Parity] Given an input sequence of values in $\{0,1\}$, output $\mod(\sum_{l=1}^Lx_l,2)$. 
    \item[Multiple-Query Associative Recall (MQAR)] Given an input sequence of key-value pairs $(k,v)\in K\times V$, followed by noise interwoven with the keys, output for each key its associated value.
    \item[$k$-hops] Given an input sequence of values from a dictionary $V$, for each element in the sequence ``hop back'' and report the value following its previous occurrence in the sequence. Keep on hopping back $k$ times. The special case of 1-hop is also known as \emph{induction heads}.
\end{description}

\paragraph{Models Description} For all models we train a single layer. The models under test are structured as: embedding layer with size matching the task vocabulary size, RMSNorm normalization layer, attention block / SSM / RNN cell, RMSNorm normalization layer, and a linear layer with size matching the task vocabulary size. We set the model dimension across all models to 64, and for Mamba2, \GRUD, and \LSTMD the hidden state dimension to 64. All models are configured with 4 heads for the attention, SSM and RNN cell respectively, and for models that use convolutional layers (Mamba2, \GRUD and \LSTMD for MQAR and $k$-hop) the convolutional kernel size is set to 4.
For Keep-n\textsuperscript{th} and Parity described below, the SSM and RNN cell contain \textit{only} the mixer (in the tables we highlight this with $^\dagger$), that is we remove all convolutional layers, normalization layers, and gating mechanisms. We do this to test the expressivity of the mixer themselves which is the main difference between the models. On the other hand, for MQAR and k-hops we use the full model (as described in  \cref{app::llm} and \cref{fig::RNN_model_schematics}) since neither Mamba2 nor \GRUD nor \LSTMD are able to solve the task without these additional components. Moreover, for Keep-n\textsuperscript{th} we add positional encoding to Mamba2, \GRUD, and \LSTMD, since it is not possible to solve the task without it (the Transformer includes positional encoding in all tasks).

\paragraph{Training Setup} For every task, we randomly obtain 10k samples as the training set and 100k samples as the test set (which we report on). For all tasks and models the sequence length is fixed to $L=100$. We train each architecture for up to 3,000 epochs with a batch size of 16. For optimization, we employ an AdamW optimizer ($\beta_1=0.9$, $\beta_2=0.999$), cosine-scheduled learning rate starting at $5\times10^-4$, and weight decay of $10^{-6}$. Tasks are trained on a single NVIDIA A100 GPU, across three different seeds (we report on the best result).
The norm of $\bb{a}_\star$ and $\bb{c}_\star$ in \GRUD and \LSTMD is clipped to a maximum of $0.90$, except for Parity, where we found that norm clipping was detrimental to the cells' ability to solve the task.

\paragraph{Additional Results} \Cref{tab::full_synthetic} expands upon the results in \cref{tab::synthetic}, by including the Keep-n\textsuperscript{th} task and k-hop for $k=1$. Overall, the observations from the main text are confirmed: nonlinear RNNs can be more effective at solving specific tasks, where the inclusion of more complex recurrence formulations aids solution.

\begin{table}[h!tb]
\footnotesize
\centering
\caption{Single-layer accuracy on synthetic tasks. Reporting best accuracy of 3 runs. $()^\dagger$ denotes results computed on the recurrent cell only. We indicate the number of parameters for each model in parentheses.}
\label{tab::full_synthetic}
\resizebox{\textwidth}{!}{
\begin{tabular}{l *{5}{c@{\hspace{12pt}}}}
    \toprule
    Model & 
        Keep-n\textsuperscript{th}$^\dagger$ & 
        Parity$^\dagger$ & 
        MQAR & 
        $k$-hop ($k=1$) & 
        $k$-hop ($k=2$) 
    \\
    & 
        $n=5, |V|=128$ & 
        $|V|=2$ & 
        $\kappa=2$, $|V|=128$ & 
        $|V|=10$ & 
        $|V|=5$
    \\
    \midrule
    Transformer & 
        100\% (48.1k) &  
         53\% (31.8k) &  
        100\% (48.1k) &  
         82\% (32.8k) &  
         78\% (32.2k)    
    \\
    Mamba2 & 
          1\% (30.1k) &  
         52\% (14.1k) &  
        100\% (38.4k) &  
        100\% (23.3k) &  
         98\% (22.7k)    
    \\
    \GRUD &
        100\% (19.8k) &  
        100\% (3.8k)  &  
        100\% (32.7k) &  
        100\% (17.6k) &  
        100\% (17.0k)    
    \\
    \LSTMD  &
        100\% (20.0k) &  
        100\% (4.0k)  &  
        100\% (32.8k) &  
        100\% (17.7k) &  
        100\% (17.1k)    
    \\
    \bottomrule
\end{tabular}}
\end{table}

\FloatBarrier
\clearpage
\section{\method software package overview}
\label{app::code}
In this section we provide an overview of the \method software package developed in the course of this research work. In its design we focused on three principles: modularity, ease of use, and extensibility. Consequently, the code separates the definition of the RNN cell and the solver, and drastically reduces the amount of code needed to be implemented by practitioners. Additionally, we provide multiple hooks for extension and generalization of various RNN cells.


\subsection{Main classes}
The core interface of the package revolves around three main classes, respectively responsible for providing abstractions to an RNN Cell, the implementation of its action, and the desired method for its application. More in detail:

\paragraph{RNN Cell} The \texttt{BaseRNNCell} class represents the abstraction of an actual RNN cell, implemented as a torch module: all user-defined cells should inherit from it. This is the only stateful class in the package, and as such it is responsible for initializing and storing the parameters defining the cell itself. These are collected in a single \texttt{SystemParameters} dataclass, to aid interoperability with the other classes in \method by providing a standardized interface for parameter access. The actual implementation of the key RNN cell functions \emph{is kept separate} from the class definition itself, and delegated to the static class \texttt{RNNCellImpl}. \texttt{BaseRNNCell} is also responsible for coordinating code execution, invoking the correct method from \texttt{RNNCellApplication} depending on the selected RNN cell application modality.

\paragraph{RNN Cell Implementation} The base static class \texttt{RNNCellImpl} collects the implementation of the functions necessary to describe the action of an RNN cell. Its main responsibility is to implement the reccurrence step \cref{eqn::rnn_seq}. Starting from this, the class leverages PyTorch \texttt{autograd} functionalities to automatically define the necessary methods to assemble the target all-at-once system \cref{eqn::state}, and particularly the computation of residuals and Jacobians in \cref{eqn::newton_it}. For additional performance, it is possible to override the latter function, and provide an explicit definition of the Jacobians---we do this ourselves in the implementation of the \GRUD and \LSTMD cells. 
This becomes particularly relevant when said Jacobians present a specific structure, that can be leveraged for making the reduction operations involved in the solution of \cref{eqn::newton_it} more computationally tractable. To this aim, in \method we provide two specializations for \texttt{RNNCellImpl}: \texttt{RNNCellDiagImpl} and \texttt{RNNCellBlockDiagImpl}, which simplify operations for those cells whose Jacobians are respectively diagonal (like for \GRUD), or composed of $N\times N$ diagonal blocks (like for \LSTMD, where $N=2$). These specializations implement the corresponding reduction steps (\cref{alg::pcr::reduce_rhs,alg::pcr::reduce_jac} in \cref{alg::pcr}), described more in detail in \cref{app::code::solver}. Our software allows to further specialize \texttt{RNNCellImpl} to encompass additional Jacobians structures (e.g., sparse, or Circulant \citep{keller2024travelingwavesencoderecent}), but the corresponding assembly, storage, and reduction operations should be provided (see also \cref{app::code::expansion}).

\paragraph{RNN Cell Application}
The third main component of \method is \texttt{RNNCellApplication}. This collection of static classes defines the different ways in which the RNN cell can be applied, and effectively implements the forward and backward passes for \texttt{BaseRNNCell}. The most vanilla version, \texttt{RNNCellSequentialApplication}, simply evaluates the cell output by sequentially applying \cref{eqn::rnn_seq} to the input sequence: this is the classical way PyTorch RNN cells act, although in our work this is mostly used for testing and debugging, or at inference time. \texttt{RNNCellParallelApplication} implements the sequence-parallel application of the RNN cell, following the method described in \cref{sec::method}, while using only PyTorch directives: the class effectively acts as a wrapper for our implementation of Newton's method, and leverages the methods from \texttt{RNNCellImpl} to iteratively assemble and solve the linearized system \cref{eqn::newton_it} via parallel reduction. A first optimization is provided in \texttt{RNNCellParallelCUDAApplication}, where the parallel solve is handled by a custom CUDA kernel, while the system assembly is still performed in PyTorch. Finally, \texttt{RNNCellParallelFusedApplication} bypasses all PyTorch-based implementations and computes the RNN application via a fully-fused CUDA kernel (implementing Newton's iterations, system assembly, and parallel reduction, end-to-end).

\subsection{Solver implementations}
\label{app::code::solver}
In addition to a programming interface for RNN cells, the main contribution of \method consists in providing ready-to-use efficient implementations of the parallel solver outlined in \cref{sec::method}.

\subsubsection{Pure PyTorch parallel reduction}
\label{app::code::solver::torch}
Our PyTorch reference implementation is a direct adaptation of the pseudocode in \cref{alg::newton,alg::pcr}. Particularly, the parallel reduction loop reduces to
\begin{minted}{python} 
@staticmethod
def parallel_reduce(
        jac: torch.Tensor,    # (B),L,?,? -> shapes will depend on
        rhs: torch.Tensor,    # (B),L,?       specific jac structure
        reduction_step: typ.Callable
) -> torch.Tensor:
    num_steps = math.ceil(math.log2(rhs.shape[-2]))
    for step in range(num_steps):
        jac, rhs = reduction_step(jac, rhs, step)
    return rhs
\end{minted}
The \texttt{reduction\_step} method defines how to perform pairwise reduction of the equations in \cref{eqn::newton_it}: namely, it implements \cref{alg::pcr::reduce_jac,alg::pcr::reduce_rhs} of \cref{alg::pcr}, adapted to a specific structure of the cell Jacobians. This is the only cell-specific method used in the solver, and by treating it as an abstraction, we are allowing for modularity in its utilization. In particular, in \method we provide implementations of this method for dense, diagonal, and block-diagonal Jacobians, covering both the general and the specific cases analyzed in this paper (c.f. \cref{sec::LSTMD}).

\subsubsection{CUDA-accelerated parallel reduction}
\label{app::code::solver::cuda}
Each level of GPU execution has access to progressively slower but larger memory systems \citep[Sec.~5.2-5.3]{cuda2025}. \emph{Threads} can access their own registers the fastest, \emph{warps} can shuffle data between threads seamlessly, and \emph{blocks} can use dedicated shared memory paying a small overhead. Only when communication occurs at \emph{grid} level must we rely on the slow global memory of the device. In implementing our kernel for parallel reduction, we minimize overhead by closely mirroring this hierarchy: the resulting algorithm directly adapts the reduction operations to each level. On top of this, rather than blindly maximizing parallelism at every level of the hierarchy, we follow a hybrid approach interweaving both sequential and parallel reductions, inspired by similar work on solvers for tri-diagonal systems \citep{manycorealgorithms}. The action of our algorithm as it traverses the GPU hierarchy is outlined next.

\begin{figure}[t!]
\centering
\vskip 0.2in
\includegraphics[trim={60mm 60mm 60mm 60mm}, clip, width=0.9\textwidth]{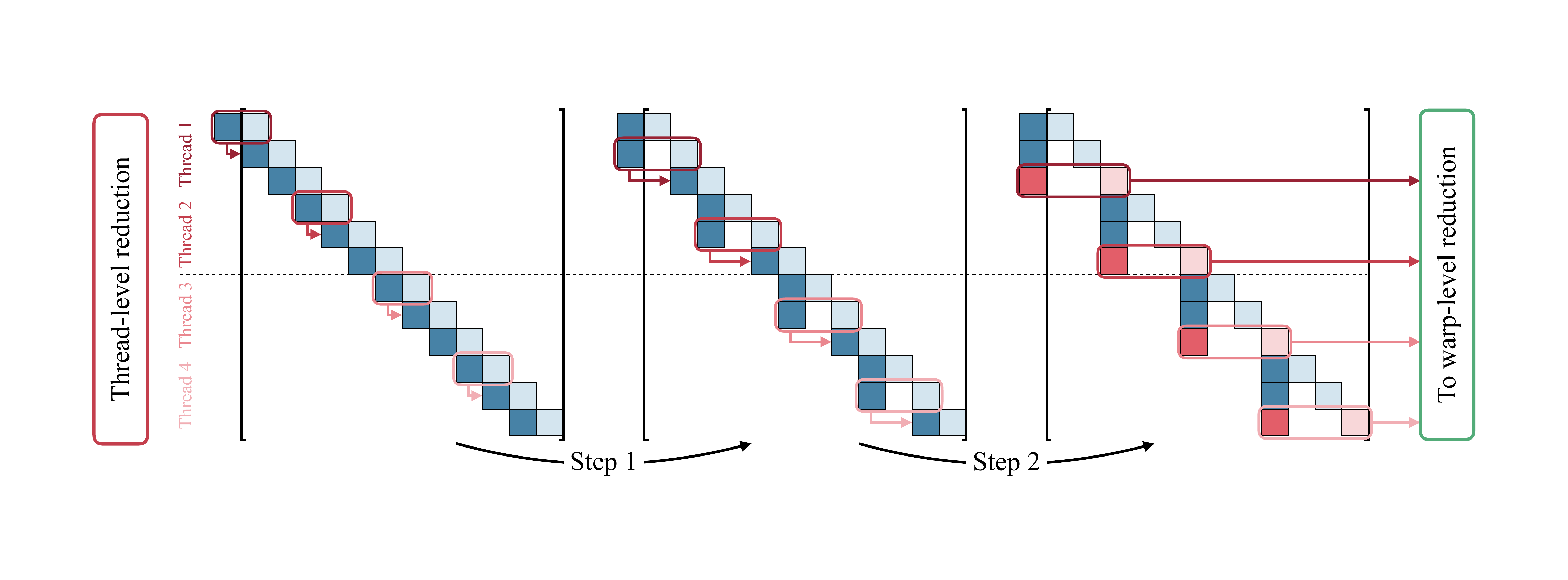}
\caption{Schematics drafting the evolution of the target block bi-diagonal system \cref{eqn::newton_it} as it undergoes the first stage of our algorithm. Light-blue squares represent its main diagonal (i.e., identity matrices), while the dark-blue once represent the Jacobians at different $l$'s.
Threads independently apply forward substitution to \texttt{chunk\_size}($=3$ in the figure) equations. At each step, this amounts to using equation $l$ to reduce the following one $l+1$ (i.e., applying \cref{alg::pcr::reduce_jac,alg::pcr::reduce_rhs} in \cref{alg::pcr}). Notice this makes all equations in a chunk depend on the value of the very first unknown, as exemplified by the progressive shift towards the left of the sub-diagonal block.
At the end of this stage, we can effectively extract a simplified system involving only the last equation of each chunk, which is then passed onto the next stage.}
\label{fig::alg_thomas}
\vskip -0.2in
\end{figure}

\paragraph{Thread-level} At the lowest hierarchy level in our algorithm, we associate each GPU thread with a small group of \texttt{chunk\_size} adjacent equations in the target system \cref{eqn::newton_it}. Threads are responsible for reading from global memory to their registers the information (namely Jacobians and residuals) associated with their chunk of equations. At this stage in the algorithm, threads must also manipulate said information and start reducing equations so that they all end up depending solely on the value of the very first unknown in their chunk. This is obtained by \emph{forward substitution}, i.e., by sequentially reducing each equation in a chunk using the previous one. Note that, while this operation is sequential, it only requires \texttt{chunk\_size}$\sim2-4$ steps, and is nonetheless conducted in parallel across threads. See \cref{fig::alg_thomas} for a diagram of the action of the algorithm at this level.

\begin{figure}[b!]
\centering
\vskip 0.2in
\includegraphics[trim={5mm 60mm 5mm 60mm}, clip, width=0.9\textwidth]{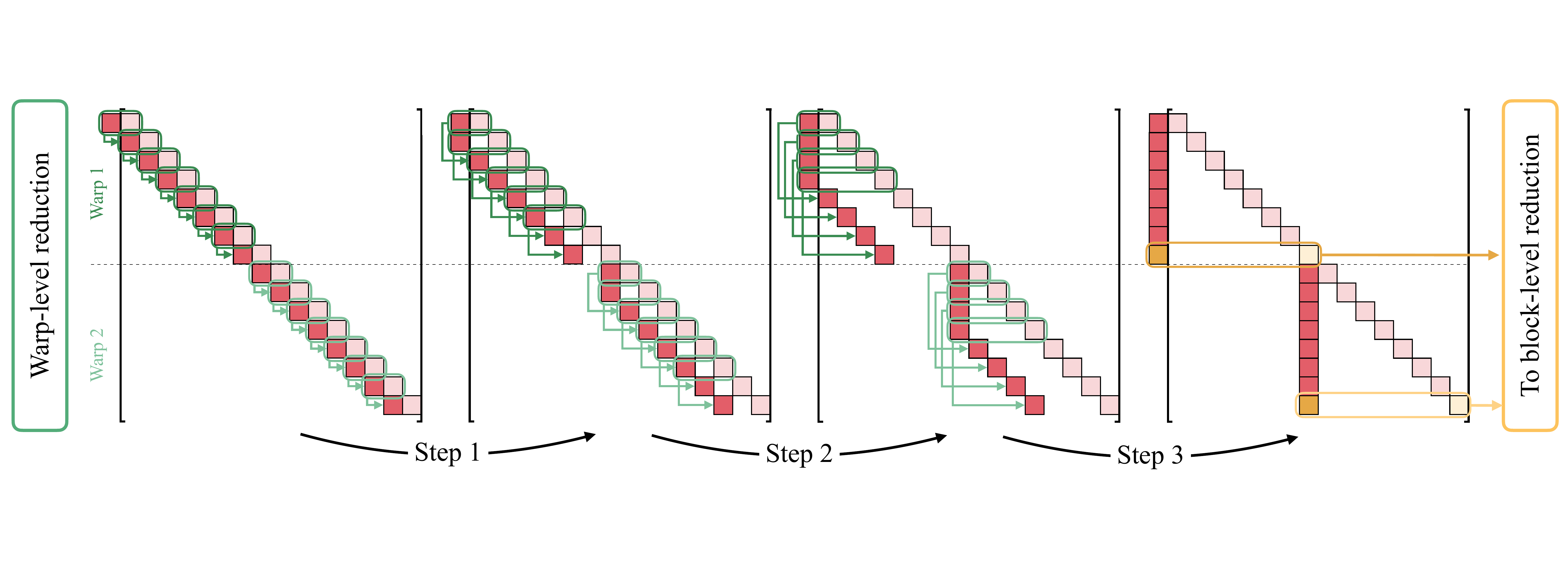}
\caption{Schematics drafting the evolution of the simplified block bi-diagonal system stemming from thread-level reduction, as it undergoes the second stage of our algorithm. The threads within a same warp (8 in the figure, but 32 in practice) apply parallel reduction to further simplify the target system. At each step $i$, this amounts to using equation $l$ to reduce the one $l+2^i$ away. Notice this progressively shifts the sub-diagonal blocks by $2^0, 2^1, 2^2,\dots$ positions, until all equations depend only on the very first unknown in the warp. At the end of this stage, we can effectively extract a simplified system involving only the last equation of each warp, which is then passed onto the next stage. At the block level, the algorithm proceeds similarly: the last threads of each warp perform parallel reduction, operating on the simplified system.}
\label{fig::alg_pcr}
\vskip -0.2in
\end{figure}

\paragraph{Warp-level} The first truly parallel reduction operations start occurring at warp level. Threads within a same warp share information from the last equations in their chunk and apply parallel reduction over them. At the end of this operation, each of these equations depends solely on the very first one in the warp. Explicitly leveraging this level in the hierarchy of the algorithm enables the use of warp-specific directives (such as warp shuffle instructions) for threads to efficiently access information stored in other threads registers, eliminating the need for expensive memory transfers or explicit synchronization. See \cref{fig::alg_pcr} for a diagram of the action of the algorithm at this level.

\paragraph{Block-level} Parallel reduction proceeds then in a similar fashion at block level, where the last equations of each warp are reduced so that they depend only on the very first one in the block. For this stage, we need to resort to the shared memory within the block to share information among warps, access to which must be explicitly synchronized: while this incurs some overhead, it remains significantly faster than accessing the device global memory. Once this stage is completed, we can traverse the hierarchy in reverse. Specifically, we use the first equation in the block to solve in parallel for the last equations of each warp, then use these to solve in parallel for the last equations of each chunk, and finally use these to solve in parallel for each equation within a chunk.
If a single block is large enough to cover the whole input sequence, then at this stage the solution to \cref{eqn::newton_it} is recovered, and the algorithm terminates here, by writing the solution back to global memory. If not, the next stage depends on the total sequence length. If it is smaller than a certain threshold, then the same block sequentially processes consecutive segments of the sequence, restarting the procedure above. For very long sequences, instead, the next level in the hierarchy is triggered.

\paragraph{Grid-level} 
For extremely long sequences, rather than having a single block covering sequentially the whole sequence length (while still applying parallel reduction within the block), we organize the GPU kernel grid so that multiple blocks are fired to cover the sequence in parallel, applying the same procedure listed above. In this case, however, the procedure leaves each block dependent on the value of its first unknown (whose solution is explicitly available only for the very first block, which covers the start of the sequence). An additional parallel reduction is then necessary to solve for the last equations of each block. This requires waiting for the completion of the kernel performing reduction \emph{within} a block, firing a second kernel to reduce equations \emph{across} blocks, and then a third one to perform the final substitutions within each block again once their solutions are available. The associated overhead grows accordingly, not only due to the additional kernel launches, but also because in-between kernels execution the device must be synchronized, and partial results must be written to global memory.

The size of the chunk of equations solved by each thread (\texttt{chunk\_size}), the number of threads allocated to a single block (\texttt{threads\_per\_block}), and the threshold number of sequential block-wise reductions (\texttt{max\_sequential\_steps}) are all hyperparameters defined at compile-time, which can be tuned to squeeze additional performance from the code by changing the per-thread workload and registers availability. The most optimal combination of parameters is a function of the actual GPU device in use, the underlying scalar datatype considered, the target input sequence length, and of the Jacobian structure of the RNN cell.
In \cref{fig::model_app_vs_chunksize,fig::model_app_vs_chunksize} we report timing results for a sweep on \texttt{chunk\_size}, which we used in this work to select the most optimal values in this work. The defaults provided in the code are a good starting point for a typical LLM training setup in mind, (i.e., sequences of $L=2^{9\sim12}$, \texttt{float32} scalars), on a NVIDIA A100 GPU.

\subsection{Example use}
\label{app::code::example}
In the following, we provide an example implementation of a simplified version of the \GRUD cell used in this work, as a use-case for the \method codebase.

As a first step, we need to introduce the parameters involved in the cell definition itself:

\begin{minted}{python} 
@dataclass
class GRUDiagSystemParameters(SystemParameters):
    A: torch.Tensor     # combines (Az,Ar,Ac)
    B: torch.Tensor     # combines (Bz,Br,Bc)
    b: torch.Tensor     # combines (bz,br,bc)
    nonlin_update: typ.Callable
    nonlin_reset: typ.Callable
    nonlin_state: typ.Callable
\end{minted}
These will be stored and initialized inside our new cell:
\begin{minted}{python} 
class GRUDiag(BaseRNNCell[
    GRUDiagConfig, GRUDiagSystemParameters, GRUDiagImpl
]):
    # invoked by base constructor
    def _specific_init(self, config: GRUDiagConfig):
        def __init__(self, config: GRUDiagConfig):
            super().__init__(config)

        # System parameters
        # - collated parameters for computing z,r,c
        kwargs = {"device": self.device, "dtype": self.dtype}
        self.A = nn.Parameter(torch.empty([3, self.state_dim], **kwargs))
        self.B = nn.Parameter(torch.empty(
            [3, self.state_dim, self.input_dim],**kwargs)
        )
        self.b = nn.Parameter(torch.empty([3, self.state_dim], **kwargs))
        self.nonlin_update = self._set_nonlinearity(config.nonlin_update)
        self.nonlin_reset = self._set_nonlinearity(config.nonlin_reset)
        self.nonlin_state = self._set_nonlinearity(config.nonlin_state)

        # Parameter initialisation
        self.reset_parameters()

    @property
    def _system_parameters(self):
        # - handily collect system parameters
        return GRUDiagSystemParameters(
            A=self.A, B=self.B, b=self.b,
            nonlin_update=self.nonlin_update,
            nonlin_reset=self.nonlin_reset,
            nonlin_state=self.nonlin_state,
        )
\end{minted}
The final ingredient to be able to start using the class, is prescribing its action \cref{eqn::GRU}:
\begin{minted}{python} 
class GRUDiagImpl(RNNCellDiagImpl[GRUDiagSystemParameters]):
    @staticmethod
    def recurrence_step(
            x: torch.Tensor,    # (B), (L), Din
            h: torch.Tensor,    # (B), (L), Dh
            system_parameters: GRUDiagSystemParameters
    ) -> torch.Tensor:
        Bxpb = torch.einsum(
            '...j,vij->...vi',
            (x, system_parameters.B)
        ) + system_parameters.b
        AhpBxpb = torch.einsum(
            'vj,...j->...vj',
            (system_parameters.A[:2, :], h)
        ) + Bxpb[..., :2, :],
        z, r = torch.unbind(AhpBxpb,dim=-2)
        z = system_parameters.nonlin_update(z)
        r = system_parameters.nonlin_reset(r)
        c = system_parameters.nonlin_state(
            system_parameters.A[2, :] * h * r + Bxpb[..., 2, :]
        )
        return (1 - z) * h + z * c
\end{minted}
This is the minimum implementation the user must provide to be able to apply the RNN cell. Just with this, the user can already toggle between sequential and parallel application\footnote{Notice that the \texttt{recurrence\_step} method is designed to work regardless of whether the input is batched or not, and whether it is a full sequence or a single element: this allows to reuse the exact same method for the different application modes of the RNN cell.} by selecting \texttt{model.mode=SEQUENTIAL} or \texttt{model.mode=PARALLEL}. Moreover, since the \GRUD cell specified above presents a diagonal structure, the user can directly leverage the efficient implementations provided in \texttt{RNNCellDiagImpl} for Jacobians assembly (via \texttt{autograd}) and manipulation, and the corresponding CUDA-accelerated solver is also available via selecting \texttt{model.mode=PARALLEL\_CUDA}. Still, by manually providing the formula for the Jacobians \cref{eqn::GRU_LSTM_jac}, one can boost performance by avoiding the reliance on \texttt{autograd}:
\begin{minted}{python} 
class GRUDiagImpl(RNNCellDiagImpl[GRUDiagSystemParameters]):
    @classmethod
    def compute_jacobians(
            cls,
            h: torch.Tensor,
            x: torch.Tensor,
            system_parameters: GRUDiagMHSystemParameters,
    ) -> torch.Tensor:
        hm1 = cls._roll_state(h)        # shift by 1 along L
        Bxpb = torch.einsum(
            '...j,vij->...vi',
            (x, system_parameters.B)
        ) + system_parameters.b
        pre_nl_z, pre_nl_r = torch.unbind(
            torch.einsum(
                'vj,...j->...vj',
                (system_parameters.A[:2, :], hm1
            )) + Bxpb[..., :2, :],
            dim=-2
        )
        z = system_parameters.nonlin_update(pre_nl_z)
        r = system_parameters.nonlin_reset(pre_nl_r)
        pre_nl_c = system_parameters.A[2, :] * hm1 * r + Bxpb[..., 2, :]
        c = system_parameters.nonlin_state(pre_nl_c)
        
        grad_z = system_parameters.derivative_nonlin_update(pre_nl_z)
        grad_r = system_parameters.derivative_nonlin_reset(pre_nl_r)
        grad_c = system_parameters.derivative_nonlin_state(pre_nl_c)
        
        J_z, J_r, J_c = torch.unbind(
            system_parameters.A * torch.stack(
                [grad_z, grad_r, grad_c], dim=-2
            ), dim=-2
        )
        J_c = J_c * (r + hm1 * J_r)

        jac = (1 - z) + (h - hm1) * J_z + z * J_c
        
        return - jac
\end{minted}

Finally, to be able to use the most optimized application mode \texttt{model.mode=PARALLEL\_FUSED}, it suffices for the user to provide equivalent CUDA implementations for the functions above. In particular:
\begin{minted}{cuda} 
template<typename scalar_t> // Implements CRTP for static polymorphism
class GRUCellDiagImpl : public RNNCellDiagImpl<
    scalar_t, GRUCellDiagImpl< scalar_t >
> {
public:
    __device__ static void readDataFromGlobal(
        const scalar_t*... glbVars,
        lclVars_t& lclVars
    ){
        // Read torch tensors in system_parameters (glbVars) and store
        //  thread-specific info in registers (lclVars)
        ...
    }
    
    __device__ static void recurrenceStep(
        const rhs_t &h,
        const lclVars_t& lclVars,
        rhs_t &hp1
    ){
        auto& [a, Bxpb] = lclVars;
        scalar_t z = nonlinUpd(   a[0] * h     + Bxpb[0] );
        scalar_t r = nonlinRes(   a[1] * h     + Bxpb[1] );
        scalar_t c = nonlinState( a[2] * h * r + Bxpb[2] );
        hp1 = (1.-z) * h + z * c;
        return;
    }

    __device__ static void computeJacobians(
        const rhs_t &h,
        const rhs_t &hm1,
        const lclVars_t& lclVars,
        jac_t &jac
    ){
        auto& [a, Bxpb] = lclVars;
        scalar_t z = a[0] * hm1 + Bxpb[0];
        scalar_t r = a[1] * hm1 + Bxpb[1];
        scalar_t jz = a[0] * derivativeNonlinUpd(z);
        scalar_t jr = a[1] * derivativeNonlinRes(r);
        z = nonlinUpd(z);
        r = nonlinRes(r);
        scalar_t c =  a[2] * hm1 * r + Bxpb[2];
        scalar_t jc = a[2] * derivativeNonlinState(c) * ( r + hm1 * jr );
        c = nonlinState(c);
        jac = - ( (1. - z) + (c - hm1) * jz + z * jc);
        return;
    }
};
\end{minted}

\subsection{Extensions to other Jacobians structures}
\label{app::code::expansion}
For RNNs with Jacobians structures different from (block-)diagonal, the user has two choices: either fall-back to the implementation for generic dense Jacobians (the base \texttt{RNNCellImpl}), or provide their own specialization. This in particular requires defining two core methods:
\begin{minted}{python} 
class MyStructuredCellImpl(
    RNNCellImpl[SystemParametersT], typ.Generic[SystemParametersT]
):
    @classmethod
    def compute_jacobians(
            cls,
            h: torch.Tensor,
            x: torch.Tensor,
            system_parameters: SystemParametersT,
    ) -> torch.Tensor:
        ...

    @classmethod
    def parallel_solve(cls) -> typ.Callable:
        return partial(parallel_reduce, reduction_step=...)
\end{minted}
The method \texttt{compute\_jacobians} not only should take care of the actual assembly of the Jacobians, but also implicitly defines how they will be represented. This must be taken into account in the definition of the specific \texttt{reduction\_step} function (which also must be implemented), responsible of describing how to conduct the reduction operations (see \cref{app::code::solver::torch,alg::pcr::reduce_rhs,alg::pcr::reduce_jac} in \cref{alg::pcr}) for this specific Jacobian structure. For the largest part, the structure of our CUDA code follows a similar pointer-to-implementation design pattern, mirroring that of the PyTorch code. In principle, then, one can provide a similar specialization for the Jacobians structure also in CUDA:
\begin{minted}{cuda} 
template < typename scalar_t, typename Derived = void >
class MyStructuredCellImpl : public RNNCellBaseImpl<
    std::conditional_t< std::is_void_v< Derived >,
        MyStructuredCellImpl< scalar_t >,
        Derived
    >   // Implements CRTP for static polymorphism
> {
    // Reduce current eq using other
    __device__ static void reduceEqs(
        const jac_t& jacOther, jac_t& jac,
        const rhs_t& rhsOther, rhs_t& rhs
    ){
        ...
    }
    ...
}
\end{minted}
However, to directly employ the current CUDA implementation of parallel reduction, the Jacobians structure should be ``lean enough'' that a single thread can hold in its registers information about (at least) one full equation of \cref{eqn::newton_it}. If this is not the case (e.g., with dense Jacobians), then the provided CUDA implementation of parallel reduction cannot be employed, and an alternative must be implemented.

%% file: iclr2026/tikz_figs/fig_newton_conv.tex
\pgfplotsset{
  newtonPlotStyle/.style={
  	tick label style={font=\normalsize},
    xmin = -0.5, xmax = 10.5, ymax = 5, ymin = 1e-8*1.5,
    xtick={0,1,...,10},
    ymode=log,
    log basis y={10},
    ymajorgrids= true,
    y grid style={white},
    unbounded coords=jump,
    xlabel= {Newton iterations $N_{\text{its}}$},
    ymajorgrids= true,
    axis background/.style={fill=gray!10},
    width=240pt,
    height=140pt,
    cycle list = {{line width=1.8pt},{opacity=0.8,line width=1.35pt},{opacity=0.6,line width=.9pt},{opacity=0.4,line width=0.45pt}},    
  }
}
\pgfplotsset{
  lineStyle/.style={
    mark=*,
    mark options={solid},
    mark size=.3pt
  }
}
\begin{tikzpicture}
\begin{groupplot}[
    newtonPlotStyle,
    group style={
        group size=2 by 2,
        horizontal sep=.5cm,
        vertical sep=0.1cm,
        x descriptions at=edge bottom,
        y descriptions at=edge left,
    },
]

\nextgroupplot[
    title=\textsc{At Initialization},
    ylabel={\GRUD},
    ylabel style={at={(axis description cs:-0.15,0.5)}, anchor=south},
    xticklabels={},
    ytick pos=left
]
\addplot+[lineStyle, gruColor] table[x index=0, y index=7,  col sep=comma ]{iclr2026/data/data_newton_convergence.csv};\label{fig::newt::gru::211}
\addplot+[lineStyle, gruColor] table[x index=0, y index=6,  col sep=comma ]{iclr2026/data/data_newton_convergence.csv};\label{fig::newt::gru::210}
\addplot+[lineStyle, gruColor] table[x index=0, y index=5,  col sep=comma ]{iclr2026/data/data_newton_convergence.csv};\label{fig::newt::gru::29}
\addplot+[lineStyle, gruColor] table[x index=0, y index=4,  col sep=comma ]{iclr2026/data/data_newton_convergence.csv};\label{fig::newt::gru::28}
\nextgroupplot[
    title=\textsc{At End of Training},
    xticklabels={},
    yticklabels={},
    ytick pos=right
]
\addplot+[lineStyle, gruColor] table[x index=0, y index=14,  col sep=comma ]{iclr2026/data/data_newton_convergence.csv};
\addplot+[lineStyle, gruColor] table[x index=0, y index=13,  col sep=comma ]{iclr2026/data/data_newton_convergence.csv};
\addplot+[lineStyle, gruColor] table[x index=0, y index=12,  col sep=comma ]{iclr2026/data/data_newton_convergence.csv};
\addplot+[lineStyle, gruColor] table[x index=0, y index=11,  col sep=comma ]{iclr2026/data/data_newton_convergence.csv};
\nextgroupplot[
    ylabel={\LSTMD},
    ylabel style={at={(axis description cs:-0.15,0.5)}, anchor=south},
    ytick pos=left
]
\addplot+[lineStyle, lstmColor] table[x index=0, y index=21,  col sep=comma ]{iclr2026/data/data_newton_convergence.csv};\label{fig::newt::lstm::211}
\addplot+[lineStyle, lstmColor] table[x index=0, y index=20,  col sep=comma ]{iclr2026/data/data_newton_convergence.csv};\label{fig::newt::lstm::210}
\addplot+[lineStyle, lstmColor] table[x index=0, y index=19,  col sep=comma ]{iclr2026/data/data_newton_convergence.csv};\label{fig::newt::lstm::29}
\addplot+[lineStyle, lstmColor] table[x index=0, y index=18,  col sep=comma ]{iclr2026/data/data_newton_convergence.csv};\label{fig::newt::lstm::28}
\nextgroupplot[
    yticklabels={},
    ytick pos=right
]
\addplot+[lineStyle, lstmColor] table[x index=0, y index=28,  col sep=comma ]{iclr2026/data/data_newton_convergence.csv};
\addplot+[lineStyle, lstmColor] table[x index=0, y index=27,  col sep=comma ]{iclr2026/data/data_newton_convergence.csv};
\addplot+[lineStyle, lstmColor] table[x index=0, y index=26,  col sep=comma ]{iclr2026/data/data_newton_convergence.csv};
\addplot+[lineStyle, lstmColor] table[x index=0, y index=24,  col sep=comma ]{iclr2026/data/data_newton_convergence.csv};
\end{groupplot}
\node[rotate=90] at ($(group c1r1.west)!0.5!(group c1r2.west) + (-2cm,0)$) {Residual norm $\max_l\|\bb{h}^k_{l} - \bb{f}(\bb{h}^k_{l-1},\bb{x}_{l})\|_\infty$};
\matrix [ampersand replacement=\&,
    scale=0.6,
    fill=white,
    fill opacity=0.9,
    matrix of nodes,
    anchor=west,
    inner xsep=2pt,
    nodes={font=\tiny,text width=4mm,align=center,text height=1.4mm},
    ] at ($(group c2r1.east)$) {
        $L$      \&                           \\
        $2^{8}$  \& \ref*{fig::newt::gru::28} \\[-4pt]
        $2^{9}$  \& \ref*{fig::newt::gru::29} \\[-4pt]
        $2^{10}$ \& \ref*{fig::newt::gru::210} \\[-4pt]
        $2^{11}$ \& \ref*{fig::newt::gru::211} \\[-4pt]
    };
\matrix [ampersand replacement=\&,
    scale=0.6,
    fill=white,
    fill opacity=0.9,
    matrix of nodes,
    anchor=west,
    inner xsep=2pt,
    nodes={font=\tiny,text width=4mm,align=center,text height=1.4mm},
    ] at ($(group c2r2.east)$) {
        $L$      \&                            \\
        $2^{8}$  \& \ref*{fig::newt::lstm::28} \\[-4pt]
        $2^{9}$  \& \ref*{fig::newt::lstm::29} \\[-4pt]
        $2^{10}$ \& \ref*{fig::newt::lstm::210} \\[-4pt]
        $2^{11}$ \& \ref*{fig::newt::lstm::211} \\[-4pt]
    };
\end{tikzpicture}

%% file: iclr2026/tikz_figs/fig_model_own_vs_mamba_zoom.tex
\pgfplotsset{
  timingPlotStyle/.style={
  	tick label style={font=\normalsize},
    xmin = 2, xmax = 2^17*1.5, ymin = 0.005, ymax = 45,
    log basis y={10},
    log basis x={2},
    ymajorgrids= true,
    y grid style={white},
    unbounded coords=jump,
    xlabel= Sequence length $L$,
    ylabel= Timings [ms],
    ymajorgrids= true,
    axis background/.style={fill=gray!10},
    legend style={pos=north west,anchor=east, font=\normalsize, very thin, draw=gray},
    cycle list = {{opacity=0.4,line width=0.45pt},{opacity=0.6,line width=.9pt},{opacity=0.8,line width=1.35pt},{line width=1.8pt},{line width=1.35pt}},    
  }
}
\pgfplotsset{
  lineStyle/.style={
    mark=*,
    mark options={solid},
    mark size=.3pt
  }
}
\begin{tikzpicture}
    \begin{loglogaxis}[name=modApp, timingPlotStyle, title=\textsc{\GRUD Forward Pass}, ytick pos=left]
        \addplot+[lineStyle,gruColor]   table[x index=0, y index=1, col sep=comma ]{iclr2026/data/data_model_app_vs_seqlength.csv};\label{fig::modZoom::GRU-seq}
        \addplot+[lineStyle,gruColor]   table[x index=0, y index=2, col sep=comma ]{iclr2026/data/data_model_app_vs_seqlength.csv};\label{fig::modZoom::GRU-par}
        \addplot+[lineStyle,gruColor]   table[x index=0, y index=3, col sep=comma ]{iclr2026/data/data_model_app_vs_seqlength.csv};\label{fig::modZoom::GRU-CUDA}
        \addplot+[lineStyle,gruColor]   table[x index=0, y index=4, col sep=comma ]{iclr2026/data/data_model_app_vs_seqlength.csv};\label{fig::modZoom::GRU-fused}
        \addplot+[lineStyle,mambaColor] table[x index=0, y index=9, col sep=comma ]{iclr2026/data/data_model_app_vs_seqlength.csv};\label{fig::modZoom::Mamba}
    \end{loglogaxis}
    \matrix [ampersand replacement=\&,
            scale=0.6,
            fill=white,
            fill opacity=0.9,
            matrix of nodes,
            anchor=north west,
            inner xsep=2pt,
            shift={(10pt,-9pt)},
            nodes={font=\tiny,text width=8mm,align=center,text height=1.4mm},
           ] at (modApp.north west) {
                \hyperref[eqn::rnn_seq]{Seq}      \& \ref*{fig::modZoom::GRU-seq}\\[-4pt]
                \hyperref[imp::torch]{PyTorch}    \& \ref*{fig::modZoom::GRU-par} \\[-4pt]
                \hyperref[imp::CUDA]{CUDA}        \& \ref*{fig::modZoom::GRU-CUDA} \\[-4pt]
                \hyperref[imp::CUDA-fused]{Fused} \& \ref*{fig::modZoom::GRU-fused} \\
                Mamba                             \& \ref*{fig::modZoom::Mamba} \\[-4pt]
        };
    \begin{loglogaxis}[name=modApp2, timingPlotStyle, title=\textsc{\LSTMD Forward Pass}, ytick pos=right, at={(modApp.outer east)}, anchor=outer west] 
        \addplot+[lineStyle,lstmColor]  table[x index=0, y index=5, col sep=comma ]{iclr2026/data/data_model_app_vs_seqlength.csv};\label{fig::modZoom::LSTM-seq}
        \addplot+[lineStyle,lstmColor]  table[x index=0, y index=6, col sep=comma ]{iclr2026/data/data_model_app_vs_seqlength.csv};\label{fig::modZoom::LSTM-par}
        \addplot+[lineStyle,lstmColor]  table[x index=0, y index=7, col sep=comma ]{iclr2026/data/data_model_app_vs_seqlength.csv};\label{fig::modZoom::LSTM-CUDA}
        \addplot+[lineStyle,lstmColor]  table[x index=0, y index=8, col sep=comma ]{iclr2026/data/data_model_app_vs_seqlength.csv};\label{fig::modZoom::LSTM-fused}
        \addplot+[lineStyle,mambaColor] table[x index=0, y index=9, col sep=comma ]{iclr2026/data/data_model_app_vs_seqlength.csv};
    \end{loglogaxis}
    \matrix [ampersand replacement=\&,
            scale=0.6,
            fill=white,
            fill opacity=0.9,
            matrix of nodes,
            anchor=north west,
            inner xsep=2pt,
            shift={(10pt,-9pt)},
            nodes={font=\tiny,text width=8mm,align=center,text height=1.4mm},
           ] at (modApp2.north west) {
                \hyperref[eqn::rnn_seq]{Seq}      \& \ref*{fig::modZoom::LSTM-seq}\\[-4pt]
                \hyperref[imp::torch]{PyTorch}    \& \ref*{fig::modZoom::LSTM-par} \\[-4pt]
                \hyperref[imp::CUDA]{CUDA}        \& \ref*{fig::modZoom::LSTM-CUDA} \\[-4pt]
                \hyperref[imp::CUDA-fused]{Fused} \& \ref*{fig::modZoom::LSTM-fused} \\
                Mamba                             \& \ref*{fig::modZoom::Mamba} \\[-4pt]
        };
\end{tikzpicture}

%% file: iclr2026/tikz_figs/fig_model_own_vs_mamba_zoom_fwdbwd.tex
\pgfplotsset{
  timingPlotStyle/.style={
  	tick label style={font=\normalsize},
    xmin = 2, xmax = 2^17*1.5, ymin = 0.1, ymax = 50,
    log basis y={10},
    log basis x={2},
    ymajorgrids= true,
    y grid style={white},
    unbounded coords=jump,
    xlabel= Sequence length $L$,
    ylabel= Timings [ms],
    ymajorgrids= true,
    axis background/.style={fill=gray!10},
    legend style={pos=north west,anchor=east, font=\normalsize, very thin, draw=gray},
    cycle list = {{opacity=0.4,line width=0.45pt},{opacity=0.6,line width=.9pt},{opacity=0.8,line width=1.35pt},{line width=1.8pt},{line width=1.35pt}},    
  }
}
\pgfplotsset{
  lineStyle/.style={
    mark=*,
    mark options={solid},
    mark size=.3pt
  }
}
\begin{tikzpicture}
    \begin{loglogaxis}[name=modApp, timingPlotStyle, title=\textsc{\GRUD Forward+Backward Pass}, ytick pos=left]
        \addplot+[lineStyle,gruColor]   table[x index=0, y index=10, col sep=comma ]{iclr2026/data/data_model_app_vs_seqlength2.csv};\label{fig::fwdbwd::GRU-seq}
        \addplot+[lineStyle,gruColor]   table[x index=0, y index=11, col sep=comma ]{iclr2026/data/data_model_app_vs_seqlength2.csv};\label{fig::fwdbwd::GRU-par}
        \addplot+[lineStyle,gruColor]   table[x index=0, y index=12, col sep=comma ]{iclr2026/data/data_model_app_vs_seqlength2.csv};\label{fig::fwdbwd::GRU-CUDA}
        \addplot+[lineStyle,gruColor]   table[x index=0, y index=13, col sep=comma ]{iclr2026/data/data_model_app_vs_seqlength2.csv};\label{fig::fwdbwd::GRU-fused}
        \addplot+[lineStyle,mambaColor] table[x index=0, y index=18, col sep=comma ]{iclr2026/data/data_model_app_vs_seqlength2.csv};\label{fig::fwdbwd::Mamba}
    \end{loglogaxis}
    \matrix [ampersand replacement=\&,
            scale=0.6,
            fill=white,
            fill opacity=0.9,
            matrix of nodes,
            anchor=north west,
            inner xsep=2pt,
            shift={(10pt,-9pt)},
            nodes={font=\tiny,text width=8mm,align=center,text height=1.4mm},
           ] at (modApp.north west) {
                \hyperref[eqn::rnn_seq]{Seq}      \& \ref*{fig::fwdbwd::GRU-seq}\\[-4pt]
                \hyperref[imp::torch]{PyTorch}    \& \ref*{fig::fwdbwd::GRU-par} \\[-4pt]
                \hyperref[imp::CUDA]{CUDA}        \& \ref*{fig::fwdbwd::GRU-CUDA} \\[-4pt]
                \hyperref[imp::CUDA-fused]{Fused} \& \ref*{fig::fwdbwd::GRU-fused} \\
                Mamba                             \& \ref*{fig::fwdbwd::Mamba} \\[-4pt]
        };
    \begin{loglogaxis}[name=modApp2, timingPlotStyle, title=\textsc{\LSTMD Forward+Backward Pass}, ytick pos=right, at={(modApp.outer east)}, anchor=outer west] 
        \addplot+[lineStyle,lstmColor]  table[x index=0, y index=14, col sep=comma ]{iclr2026/data/data_model_app_vs_seqlength2.csv};\label{fig::fwdbwd::LSTM-seq}
        \addplot+[lineStyle,lstmColor]  table[x index=0, y index=15, col sep=comma ]{iclr2026/data/data_model_app_vs_seqlength2.csv};\label{fig::fwdbwd::LSTM-par}
        \addplot+[lineStyle,lstmColor]  table[x index=0, y index=16, col sep=comma ]{iclr2026/data/data_model_app_vs_seqlength2.csv};\label{fig::fwdbwd::LSTM-CUDA}
        \addplot+[lineStyle,lstmColor]  table[x index=0, y index=17, col sep=comma ]{iclr2026/data/data_model_app_vs_seqlength2.csv};\label{fig::fwdbwd::LSTM-fused}
        \addplot+[lineStyle,mambaColor] table[x index=0, y index=18, col sep=comma ]{iclr2026/data/data_model_app_vs_seqlength2.csv};
    \end{loglogaxis}
    \matrix [ampersand replacement=\&,
            scale=0.6,
            fill=white,
            fill opacity=0.9,
            matrix of nodes,
            anchor=north west,
            inner xsep=2pt,
            shift={(10pt,-9pt)},
            nodes={font=\tiny,text width=8mm,align=center,text height=1.4mm},
           ] at (modApp2.north west) {
                \hyperref[eqn::rnn_seq]{Seq}      \& \ref*{fig::fwdbwd::LSTM-seq}\\[-4pt]
                \hyperref[imp::torch]{PyTorch}    \& \ref*{fig::fwdbwd::LSTM-par} \\[-4pt]
                \hyperref[imp::CUDA]{CUDA}        \& \ref*{fig::fwdbwd::LSTM-CUDA} \\[-4pt]
                \hyperref[imp::CUDA-fused]{Fused} \& \ref*{fig::fwdbwd::LSTM-fused} \\
                Mamba                             \& \ref*{fig::fwdbwd::Mamba} \\[-4pt]
        };
\end{tikzpicture}

%% file: iclr2026/tikz_figs/fig_kernel_time_vs_chunksize.tex
\pgfplotsset{
  timingPlotStyle/.style={
  	tick label style={font=\normalsize},
    xmin = 2, xmax = 2^22*1.5, ymin = 0.005, ymax = 0.5,
    log basis y={10},
    log basis x={2},
    ymajorgrids= true,
    y grid style={white},
    unbounded coords=jump,
    xlabel= Sequence length $L$,
    ylabel= Timings [ms],
    ymajorgrids= true,
    axis background/.style={fill=gray!10},
    cycle list = {{opacity=0.33,line width=.6pt},{opacity=0.66,line width=1.2pt},{line width=1.8pt},{line width=1.2pt}},    
  }
}
\pgfplotsset{
  lineStyle/.style={
    mark=*,
    mark options={solid},
    mark size=.3pt
  }
}
\begin{tikzpicture}
    \begin{loglogaxis}[name=modApp, timingPlotStyle, title=\textsc{Diagonal}, ytick pos=left] 
        \addplot+[lineStyle,gruColor]    table[x index=0, y index=1, col sep=comma ]{iclr2026/data/data_kernel_time_vs_chunksize.csv};\label{fig::timevschunk::diag-1}
        \addplot+[lineStyle,gruColor]    table[x index=0, y index=2, col sep=comma ]{iclr2026/data/data_kernel_time_vs_chunksize.csv};\label{fig::timevschunk::diag-2}
        \addplot+[lineStyle,gruColor]    table[x index=0, y index=3, col sep=comma ]{iclr2026/data/data_kernel_time_vs_chunksize.csv};\label{fig::timevschunk::diag-4}
        \addplot+[lineStyle, mambaColor] table[x index=0, y index=5, col sep=comma ]{iclr2026/data/data_kernel_time_vs_seqlength.csv};\label{fig::timevschunk::Mamba}
    \end{loglogaxis}
    \matrix [ampersand replacement=\&,
            scale=0.6,
            fill=white,
            fill opacity=0.9,
            matrix of nodes,
            anchor=north west,
            inner xsep=2pt,
            shift={(10pt,-9pt)},
            nodes={font=\tiny,text width=7mm,align=center,text height=1.4mm},
           ] at (modApp.north west) {
                $c=1$ \& \ref*{fig::timevschunk::diag-1} \\[-4pt]
                $c=2$ \& \ref*{fig::timevschunk::diag-2} \\[-4pt]
                $c=4$ \& \ref*{fig::timevschunk::diag-4} \\
                |[text width=11mm]|Mamba scan\& \ref*{fig::timevschunk::Mamba} \\[-4pt]
        };
    \begin{loglogaxis}[name=modApp2, timingPlotStyle, title=\textsc{$2\times2$ Block-Diagonal}, ytick pos=right, at={(modApp.outer east)}, anchor=outer west] 
        \addplot+[lineStyle,lstmColor]   table[x index=0, y index=6, col sep=comma ]{iclr2026/data/data_kernel_time_vs_chunksize.csv};\label{fig::timevschunk::bdiag-1}
        \addplot+[lineStyle,lstmColor]   table[x index=0, y index=7, col sep=comma ]{iclr2026/data/data_kernel_time_vs_chunksize.csv};\label{fig::timevschunk::bdiag-2}
        \addplot+[lineStyle,lstmColor]   table[x index=0, y index=8, col sep=comma ]{iclr2026/data/data_kernel_time_vs_chunksize.csv};\label{fig::timevschunk::bdiag-4}
        \addplot+[lineStyle, mambaColor] table[x index=0, y index=5, col sep=comma ]{iclr2026/data/data_kernel_time_vs_seqlength.csv};
    \end{loglogaxis}
    \matrix [ampersand replacement=\&,
            scale=0.6,
            fill=white,
            fill opacity=0.9,
            matrix of nodes,
            anchor=north west,
            inner xsep=2pt,
            shift={(10pt,-9pt)},
            nodes={font=\tiny,text width=7mm,align=center,text height=1.4mm},
           ] at (modApp2.north west) {
                $c=1$ \& \ref*{fig::timevschunk::bdiag-1} \\[-4pt]
                $c=2$ \& \ref*{fig::timevschunk::bdiag-2} \\[-4pt]
                $c=4$ \& \ref*{fig::timevschunk::bdiag-4} \\
                |[text width=11mm]|Mamba scan\& \ref*{fig::timevschunk::Mamba} \\[-4pt]
        };
\end{tikzpicture}

%% file: iclr2026/tikz_figs/fig_model_app_vs_chunksize.tex
\pgfplotsset{
  timingPlotStyle/.style={
  	tick label style={font=\normalsize},
    xmin = 2, xmax = 2^17*1.5, ymin = 0.005, ymax = 4.5,
    log basis y={10},
    log basis x={2},
    ymajorgrids= true,
    y grid style={white},
    unbounded coords=jump,
    xlabel= Sequence length $L$,
    ylabel= Timings [ms],
    ymajorgrids= true,
    axis background/.style={fill=gray!10},
    cycle list = {{opacity=0.33,line width=.6pt},{opacity=0.66,line width=1.2pt},{line width=1.8pt},{line width=1.2pt}},    
  }
}
\pgfplotsset{
  lineStyle/.style={
    mark=*,
    mark options={solid},
    mark size=.3pt
  }
}
\begin{tikzpicture}
    \begin{loglogaxis}[name=modApp, timingPlotStyle, title=\textsc{\GRUD Forward Pass}, ytick pos=left]
        \addplot+[lineStyle,gruColor]   table[x index=0, y index=6, col sep=comma ]{iclr2026/data/data_model_app_vs_chunksize.csv};\label{fig::appvschunk::GRU-1}
        \addplot+[lineStyle,gruColor]   table[x index=0, y index=7, col sep=comma ]{iclr2026/data/data_model_app_vs_chunksize.csv};\label{fig::appvschunk::GRU-2}
        \addplot+[lineStyle,gruColor]   table[x index=0, y index=8, col sep=comma ]{iclr2026/data/data_model_app_vs_chunksize.csv};\label{fig::appvschunk::GRU-4}
        \addplot+[lineStyle,mambaColor] table[x index=0, y index=9, col sep=comma ]{iclr2026/data/data_model_app_vs_seqlength.csv};\label{fig::appvschunk::Mamba}
    \end{loglogaxis}
    \matrix [ampersand replacement=\&,
            scale=0.6,
            fill=white,
            fill opacity=0.9,
            matrix of nodes,
            anchor=north west,
            inner xsep=2pt,
            shift={(10pt,-9pt)},
            nodes={font=\tiny,text width=7mm,align=center,text height=1.4mm},
           ] at (modApp.north west) {
                $c=1$ \& \ref*{fig::appvschunk::GRU-1} \\[-4pt]
                $c=2$ \& \ref*{fig::appvschunk::GRU-2} \\[-4pt]
                $c=4$ \& \ref*{fig::appvschunk::GRU-4} \\
                Mamba \& \ref*{fig::appvschunk::Mamba} \\[-4pt]
        };
    \begin{loglogaxis}[name=modApp2, timingPlotStyle, title=\textsc{\LSTMD Forward Pass}, ytick pos=right, at={(modApp.outer east)}, anchor=outer west] 
        \addplot+[lineStyle,lstmColor]  table[x index=0, y index=16, col sep=comma ]{iclr2026/data/data_model_app_vs_chunksize.csv};\label{fig::appvschunk::LSTM-1}
        \addplot+[lineStyle,lstmColor]  table[x index=0, y index=17, col sep=comma ]{iclr2026/data/data_model_app_vs_chunksize.csv};\label{fig::appvschunk::LSTM-2}
        \addplot+[lineStyle,lstmColor]  table[x index=0, y index=18, col sep=comma ]{iclr2026/data/data_model_app_vs_chunksize.csv};\label{fig::appvschunk::LSTM-4}
        \addplot+[lineStyle,mambaColor] table[x index=0, y index=9, col sep=comma ]{iclr2026/data/data_model_app_vs_seqlength.csv};
    \end{loglogaxis}
    \matrix [ampersand replacement=\&,
            scale=0.6,
            fill=white,
            fill opacity=0.9,
            matrix of nodes,
            anchor=north west,
            inner xsep=2pt,
            shift={(10pt,-9pt)},
            nodes={font=\tiny,text width=7mm,align=center,text height=1.4mm},
           ] at (modApp2.north west) {
                $c=1$ \& \ref*{fig::appvschunk::LSTM-1} \\[-4pt]
                $c=2$ \& \ref*{fig::appvschunk::LSTM-2} \\[-4pt]
                $c=4$ \& \ref*{fig::appvschunk::LSTM-4} \\
                Mamba \& \ref*{fig::appvschunk::Mamba}  \\[-4pt]
        };
\end{tikzpicture}